\documentclass{article} \usepackage{iclr2025_conference,times}

\usepackage{amsmath,amsfonts,bm}

\def\eqref#1{(\ref{#1})}

\def\1{\bm{1}}

\usepackage{hyperref}
\usepackage{url}
\usepackage{graphicx}
\usepackage{subcaption}
\usepackage{enumitem} 
\usepackage{csquotes}
\MakeOuterQuote{"}

\usepackage{float} 

\title{Input--Label Correlation Governs\\ a Linear-to-Nonlinear Transition\\ in Random Features under Spiked Covariance}

\author{Samet Demir$^{1}$ \& Zafer Do\u{g}an$^{1,2}$ \\
$^1$Machine Learning and Information Processing Group, KUIS AI Center\\
$^2$Department of Electrical and Electronics Engineering\\
Ko\c{c} University, İstanbul, Turkey\\
\texttt{\{sdemir20,zdogan\}@ku.edu.tr}
}

\iclrfinalcopy 

\usepackage{amsmath}
\usepackage{amssymb}
\usepackage{mathtools}
\usepackage{amsthm}
\usepackage{enumitem} 
\usepackage{float}
\usepackage{dsfont}

\theoremstyle{plain}
\newtheorem{theorem}{Theorem}

\newtheorem{lemma}[theorem]{Lemma}
\newtheorem{corollary}[theorem]{Corollary}
\theoremstyle{definition}

\theoremstyle{remark}
\newtheorem{remark}[theorem]{Remark}

\def\Normal{{\mathcal N}}
\def\w{{\boldsymbol \omega}}
\def\F{{\mathbf F}}
\def\f{{\mathbf f}}
\def\x{{\mathbf x}}
\def\z{{\mathbf z}}

\def\R{\mathbb{R}}
\def\I{{\mathbf I}}
\def\bold1{{\mathbf 1}}

\def\bgamma{{\boldsymbol \gamma}}
\def\bxi{{\boldsymbol \xi}}

\def\g{{\mathbf g}}

\def\a{{\mathbf a}}
\def\b{{\mathbf b}}
\def\G{{\mathcal G}}
\def\T{{\mathcal T}}

\def\Prob{{\mathbb P}}
\def\A{{\mathbf A}}
\def\B{{\mathbf B}}

\def\bR{{\mathbf R}}
\def\br{{\mathbf r}}
\def\D{{\mathbf D}}

\DeclareMathOperator*{\argmin}{arg\,min}
\def\E{\mathop{{\mathbb E}}}

\begin{document}

\maketitle

\begin{abstract}
Random feature models (RFMs)---two-layer networks with a randomly initialized
fixed first layer and a trained linear readout---are among the simplest
nonlinear predictors. Prior asymptotic analyses in the proportional
high-dimensional regime show that, under isotropic data, RFMs reduce to noisy
linear models and offer no advantage over classical linear methods such as
ridge regression. Yet RFMs frequently outperform linear baselines on
structured real data. We show that this tension is explained by a
\emph{correlation-driven phase transition}: under spiked-covariance designs,
the interaction between anisotropy and input--label correlation determines
whether the RFM behaves as an effectively linear predictor or exhibits
genuinely nonlinear gains. Concretely, we establish a universality principle
under anisotropy and characterize the RFM generalization error via an
equivalent \emph{noisy polynomial} model. The effective degree of this
polynomial---equivalently, which Hermite orders of the activation
survive---is governed by the strength of input--label correlation, yielding an
explicit boundary in the correlation--spike-magnitude plane. Below the boundary,
the RFM collapses to a linear surrogate and can underperform strong linear
baselines; above it, higher-order terms persist and the RFM achieves a clear
nonlinear advantage. Numerical simulations and real-data experiments corroborate the
theory and delineate the transition between these two regimes.
\end{abstract}

\section{Introduction}
\label{sec:intro}
Random feature models (RFMs)~\cite{rahimi2007random}---two-layer networks with a randomly initialized and fixed first layer and a trained linear readout---are among the simplest nonlinear predictors and a canonical proxy for studying neural-network-style representations without feature learning. A recurring question in modern high-dimensional statistics and learning theory is whether, and when, the nonlinear representation induced by a random first layer yields a \emph{genuine} improvement over classical linear prediction in high dimensions.

\begin{figure*}
    \centering
    \begin{subfigure}[b]{0.4\textwidth}
         \centering
         \includegraphics[width=0.99\linewidth]{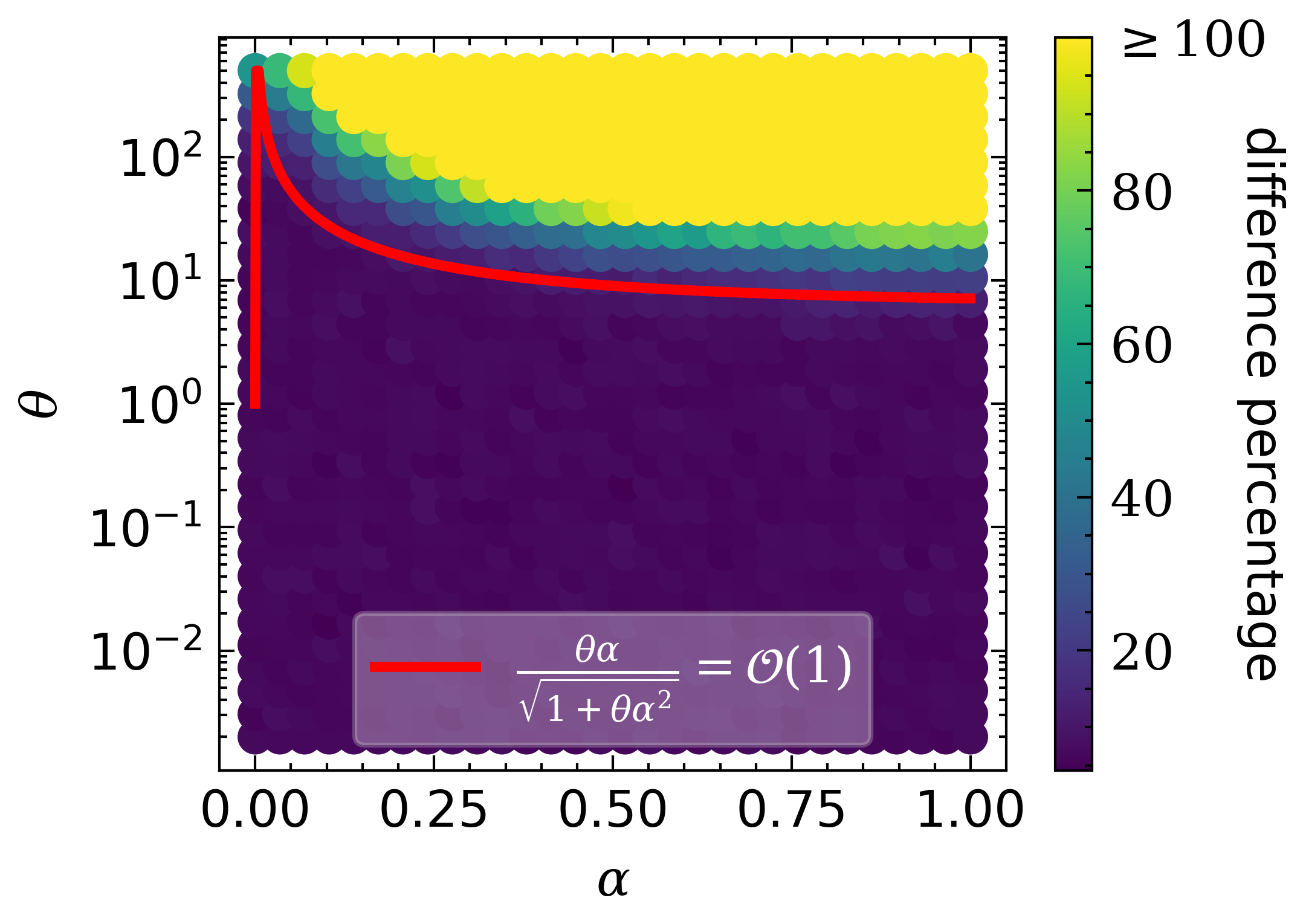}
         \caption{Phase diagram: percentage difference between generalization errors of the RFM and the noisy linear model~\eqref{eq:gaussian_model} across alignment $\alpha$ and spike magnitude $\theta$.}
         \label{fig:phase_diagram}
     \end{subfigure}
     \hspace{2em}
     \begin{subfigure}[b]{0.35\textwidth}
         \centering
         \includegraphics[width=0.99\linewidth]{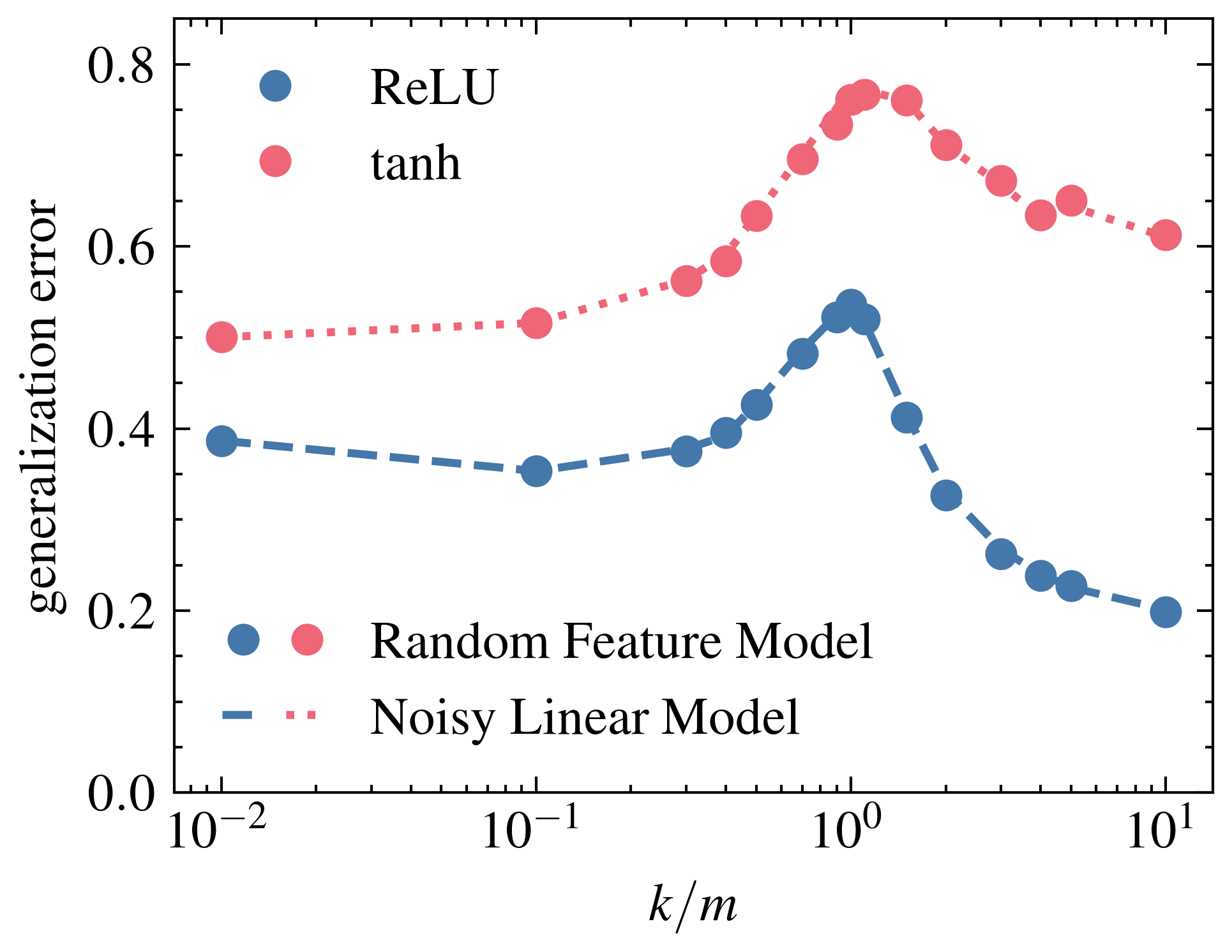}
         \caption{Linear regime:  generalization errors of the RFM and the noisy linear model~\eqref{eq:gaussian_model} for $\alpha = \mathcal{O}(1/\sqrt{n})$ \textbf{(misaligned)} and $\theta = n^{1/2}$.}
        \label{fig:linear_equivalence}
     \end{subfigure}
     \caption{\textit{The phase boundary in the $(\alpha,\theta)$~plane.} Left: the red curve separates the effectively linear regime (below) from the genuinely nonlinear regime (above). The red curve marks the phase boundary predicted by Corollary~\ref{cor:linear}. Right: when $\alpha = \mathcal{O}(1/\sqrt{n})$, the RFM lies in the linear regime and matches the noisy linear model across activations. Here $\sigma_*=\sigma_{ReLU}$, $\lambda=10^{-2}$, $n=400$, $m=500$; averages over 50 Monte Carlo runs. Note that we provide our simulation results for the nonlinear regime in Figure \ref{fig:polynomial_equivalence}.}
\end{figure*}

A substantial body of recent work provides sharp asymptotic characterizations of RFMs in the \emph{proportional} high-dimensional regime, where the sample size $m$, dimension $n$, and width $k$ diverge jointly with fixed ratios. Under \emph{isotropic} covariates, these analyses show that RFMs reduce to appropriately matched \emph{noisy linear} models~\cite{montanari2019generalization,gerace2020generalisation,goldt2022gaussian,mei2022generalization,hu2022universality}, implying that random nonlinear features provide no advantage over linear prediction. Yet RFMs frequently outperform linear baselines on structured real data. This paper explains the tension:
\begin{quote}
\emph{Under spiked-covariance data, a \textbf{correlation-driven phase transition} separates an effectively linear regime from a genuinely nonlinear one. The transition boundary is governed by the anisotropy and input--label correlation.}
\end{quote}
Below the boundary, the RFM collapses to a noisy linear predictor and can underperform strong linear baselines; above it, higher-order polynomial terms in the activation persist and the RFM achieves a clear nonlinear advantage. Figure~\ref{fig:phase_diagram} displays this phase diagram in the correlation--spike-magnitude plane.

\paragraph{Random feature regression in the proportional regime}
We study the standard supervised learning setup with an RFM predictor
\begin{equation}
    \label{eq:feature_model}
    f_\sigma(\x) \;=\; \w^T \sigma(\F \x),
\end{equation}
where $\x\in\R^n$ is the input, $\F\in\R^{k\times n}$ is a random feature matrix sampled at initialization and then held fixed, $\sigma:\R\to\R$ is an activation applied entrywise, and $\w\in\R^k$ is learned from data. Given training samples $\{(\x_i,y_i)\}_{i=1}^m$, we learn $\w$ by ridge-regularized least squares,
\begin{equation}
\label{eq:train}
\hat{\w}_\sigma
\;:=\;
\operatorname*{argmin}_{\w\in\R^k}\;
\frac{1}{m}\sum_{i=1}^m \big(y_i-\w^T\sigma(\F\x_i)\big)^2 \;+\; \lambda\|\w\|_2^2,
\end{equation}
with corresponding training error $\T_{\sigma}$ and evaluate performance via the generalization error
\begin{equation}
\label{eq:generalization}
\G_\sigma
\;:=\;
\E_{(\x,y)}\big(y-\hat{\w}_\sigma^T\sigma(\F\x)\big)^2.
\end{equation}
Our focus is the proportional asymptotic limit $m,n,k\to\infty$ with $n/m,n/k\in(0,\infty)$, which is also called the \emph{linear scaling} regime.

\paragraph{Why isotropic theory predicts ``effective linearity''}
Under isotropic covariates (e.g., $\x\sim\Normal(0,\I_n)$), RFMs admit a precise asymptotic reduction to a matched noisy linear model~\cite{montanari2019generalization,gerace2020generalisation,goldt2022gaussian,mei2022generalization,hu2022universality}. In particular, the RFM is asymptotically equivalent (in performance) to
\begin{align}
\label{eq:gaussian_model}
\w^T(\mu_0\boldsymbol{1}+\mu_1\F\x+\mu_*\z),
\end{align}
where $\boldsymbol{1}$ is the all-ones vector, $\z\sim\Normal(0,\I_k)$ is independent noise, and the constants $\mu_0,\mu_1,\mu_*$ depend on $\sigma$ via
$\mu_0=\E[\sigma(z)]$, $\mu_1=\E[z\sigma(z)]$, and
$\mu_*=\sqrt{\E[\sigma(z)^2]-\mu_0^2-\mu_1^2}$ for $z\sim\Normal(0,1)$.
This reduction underpins sharp predictions for training/generalization curves and double-descent behavior across losses, activations, and regularization~\cite{dhifallah2020precise,Belkin19,Geiger_2020,mei2022generalization,wang2022optimal,demir2023optimal,nakkiran2020optimal}. Crucially, it also implies that, in isotropic settings, random nonlinear features do \emph{not} yield gains beyond an appropriately matched linear predictor---a conclusion that holds regardless of the choice of activation.

\paragraph{Spiked covariance and input--label correlation}
The isotropic reduction~\eqref{eq:gaussian_model} is tightly coupled to rotational symmetry of the input distribution. Real covariates are rarely isotropic; they often exhibit low-dimensional structure, anisotropic variance profiles, and alignments between directions of large variance and directions predictive of the label~\cite{facco2017estimating}. We study RFMs under a structured but analytically tractable anisotropic model: spiked-covariance covariates with a single-index nonlinear teacher~\cite{spikedcovariance1,spikedcovariance2},
\begin{align}
\label{eq:xdata}
\x &\sim \Normal(0,\I_n+\theta\,\bgamma\bgamma^T),
\qquad
y:=\sigma_*\!\left(\frac{\bxi^T \x}{\sqrt{1+\theta\alpha^2}}\right),
\end{align}
where $\bgamma\in\R^n$ is a fixed ``spike'' direction with $\|\bgamma\|_2=1$, $\bxi\in\R^n$ is a fixed ``label'' direction with $\|\bxi\|_2=1$, and $\alpha := \bgamma^T\bxi$ is the \emph{input--label correlation} (alignment) parameter. The spike magnitude $\theta$ scales as $\theta\asymp n^\beta$ for $\beta\in[0,1/2)$, and the normalization in~\eqref{eq:xdata} ensures the teacher input has unit variance. This data model isolates, in the simplest possible form, the interaction between (i)~\emph{anisotropy} in the covariates and (ii)~\emph{correlation} between the anisotropic direction and the label, which together determine on which side of the phase boundary the RFM operates. This minimal model isolates the mechanism and yields predictions that remain qualitatively robust under real-world data (see Figure~\ref{fig:CIFAR-10}). Our analysis extends naturally to multi-spike covariance
$\Sigma = \I_n + \sum_{r=1}^{R} \theta_r \bgamma_r \bgamma_r^T$
with a multi-index teacher
$y = \sigma_*(\boldsymbol{\Xi}^T\x)$ for $\boldsymbol{\Xi} \in \R^{n \times R}$: the
quantity $\eta$ generalizes to a vector of per-spike correlations,
and the phase boundary becomes a surface in the joint
$(\alpha_1, \theta_1, \ldots, \alpha_R, \theta_R)$ space. We
focus on the rank-one case ($R=1$) throughout, as it already
captures the core mechanism---the $\theta\alpha$ amplification in
the cross-covariance---while keeping the phase diagram
two-dimensional and the exposition transparent.

\paragraph{The phase transition and its mechanism}
Our analysis reveals an explicit transition in the effective nonlinearity of the RFM, controlled by a single quantity $\eta$ that couples the input--label correlation $\alpha$, the spike magnitude $\theta$, and the feature matrix $\F$ (see~\eqref{eq:eta}). The mechanism is as follows. The activation $\sigma$ admits a Hermite expansion whose terms contribute to prediction at different polynomial orders. Under isotropy, all higher-order contributions are suppressed and only the linear term survives---recovering the classical noisy linear surrogate. Under spiked covariance, the quantity $\eta$ modulates which Hermite orders persist: when $\eta = \mathcal{O}(n^{-1/2})$, only the linear term contributes and the RFM remains effectively linear; when $\eta$ is larger, higher-order terms activate and the RFM becomes a genuinely nonlinear predictor. The boundary $\eta = \mathcal{O}(n^{-1/2})$ traces an explicit curve in the $(\alpha, \theta)$~plane (Figure~\ref{fig:phase_diagram}), cleanly separating the two regimes.

\paragraph{Contributions}
Our contributions are:
\begin{enumerate}
    \item \textbf{A correlation-driven phase transition in effective nonlinearity.}
    We identify an explicit boundary in the correlation--spike-magnitude plane that separates an effectively linear regime from a genuinely nonlinear one. Below the boundary, the RFM collapses to the classical noisy linear surrogate and may underperform strong linear baselines; above it, the RFM achieves strictly smaller generalization error through persistent higher-order terms. This provides a data-centric criterion for when random nonlinear features are truly useful in the proportional regime.

    \item \textbf{Noisy polynomial surrogate with correlation-controlled effective degree.}
    We characterize the RFM generalization error via an equivalent noisy polynomial model whose degree---equivalently, which Hermite orders of the activation survive---is governed by the input--label correlation. This generalizes the classical noisy linear reduction~\eqref{eq:gaussian_model} and makes the role of higher-order effects explicit.

    \item \textbf{Universality under anisotropy.}
    As a key technical tool, we establish an activation-level universality theorem under the spiked covariance model: RFM performance depends on the activation only through certain low-order joint moments of $(\sigma(\F\x),y)$. This extends the universality framework of~\cite{hu2022universality} beyond isotropic designs.
\end{enumerate}
Our theoretical predictions are supported by experiments, confirming the predicted phase boundary and the transition between effectively linear and genuinely nonlinear behavior.

\section{Related Work}
\label{sec:related_work}

\paragraph{Random features, kernels, and linearized network models}
Random feature models (RFMs) were introduced as randomized approximations to kernel methods~\cite{rahimi2007random}. They have since become a canonical object in learning theory, also because RFMs and the Neural Tangent Kernel (NTK) provide complementary linearized viewpoints of two-layer neural networks~\cite{ghorbani2021linearized}. Although fully trained networks can empirically outperform both RFMs and NTK predictors~\cite{ghorbani2020neural}, RFMs remain a key testbed for isolating the role of a fixed nonlinear map from that of feature learning.

\paragraph{Proportional-regime asymptotics and noisy-linear equivalence}
A large literature gives sharp characterizations of RFM training and generalization in the proportional regime under isotropic covariates, via an asymptotic reduction to a matched noisy linear surrogate~\cite{montanari2019generalization,gerace2020generalisation,goldt2020modeling,goldt2022gaussian,dhifallah2020precise,mei2022generalization}. In particular, \cite{hu2022universality} proves such an equivalence via Lindeberg-type arguments, and \cite{montanari2022universality} establishes related universality for empirical risk minimization with Gaussian feature replacements. These results underpin precise learning-curve predictions and double-descent analyses~\cite{dhifallah2020precise,mei2022generalization}, but they are confined to rotationally invariant data and therefore predict that random nonlinear features offer no advantage over linear prediction. They do not identify structured data settings where this collapse breaks, nor provide an explicit criterion for the transition between linear and nonlinear behavior---the focus of our work.

\paragraph{Beyond isotropy: failures of Gaussian equivalence}
Recent works show that Gaussian equivalence need not persist under structured data~\cite{gerace2022gaussian,pesce2023gaussian, mai2025the}, motivating alternative representations such as Gaussian mixtures~\cite{dandi2023universality, demir2025asymptotic}. Our work addresses a complementary question: rather than seeking alternative feature replacements, we characterize \emph{what replaces} the linear surrogate when it breaks under anisotropy. We introduce a concrete noisy-polynomial surrogate for RF regression and show that its \emph{effective degree} (equivalently, which Hermite orders survive) is controlled by input--label correlation, yielding an explicit correlation-driven boundary separating effectively linear and genuinely nonlinear regimes.

\paragraph{Feature learning}
A parallel line of work studies minimal feature learning via one or few gradient steps, often obtaining improvements over fixed random features~\cite{ba2022highdimensional,damian2022neural}. \cite{ba2023learning,mousavi2024gradient} analyze sample-complexity benefits under spiked-covariance designs, while \cite{moniri2023theory,dandi2023learning,cui2024asymptotics} characterize the spectral spike induced by gradient updates and establish Gaussian-equivalence statements under isotropic data. \cite{demir2025asymptotic} extends these Gaussian-equivalence statements to anisotropic mixture data. These analyses exploit data structure through updates to the first layer. Our result is complementary: we show that even without feature learning, anisotropy together with input--label correlation can induce a phase transition in the effective behavior of fixed random features, delineating the baseline that feature-learning methods must improve upon.

\paragraph{Deep random features and alternative scalings}
Deep random features and anisotropic random layers have been analyzed via precise asymptotics and statistical-physics techniques~\cite{schroder2023deterministic,bosch2023precise,zavatone2023learning}. Complementary work considers polynomial scaling limits~\cite{hu2024asymptotics, lu2025equivalence, wen2025does}, where generalization behavior changes qualitatively with the scaling exponents. These works modify the architecture or scaling regime; in contrast, we keep both fixed and show that data structure alone---anisotropy together with input--label correlation---is sufficient to trigger a linear-to-nonlinear transition in the standard proportional regime.

\section{Preliminaries}
\label{sec:preliminaries}

\paragraph{Notations}
We denote vectors and matrices using bold letters, with lowercase letters representing vectors and uppercase letters representing matrices. Scalars are indicated by non-bold letters. The notation $||\x||_2$ refers to the Euclidean norm of the vector $\x$ while $||\F||$ denotes the spectral norm of the matrix $\F$. We use the term $\text{polylog } k$ to represent any polylogarithmic function of $k$. The symbol $\stackrel{\Prob}{\to}$ indicates convergence in probability. We employ big-O notation, denoted $\mathcal{O}(.)$, and small-o notation, represented as $o(.)$ with respect to the parameters $k,n,m$. We also use the notation $\theta \asymp n^\beta$ to indicate that there exist constants $C_1, C_2 > 0$ such that $C_1 n^\beta \leq \theta \leq C_2 n^\beta$. Finally, we denote the $i$-th Hadamard power of a vector $\x$ as $\x^{\circ i}$.

\label{sec:assumptions}
\paragraph{Suppositions}
We establish our results based on the following assumptions.

\begin{enumerate}[label=S.\arabic* ---]
    \item The spike and label signal vectors $\bgamma, \bxi \in \R^n$ are deterministic and $||\bgamma||_2 = ||\bxi||_2 = 1$.
    \item $\x \sim \Normal(0, \I_n + \theta \bgamma \bgamma^T)$ for spike magnitude $\theta \asymp n^\beta$ where $\beta \in [0,1/2)$.
    \item $y := \sigma_*\left(\frac{\bxi^T \x}{\sqrt{1+\theta\alpha^2}}\right)$ where $\sigma_*: \R \to \R$ is a function satisfying $|\sigma_*(x)| < C (1 + |x|^K)$ for all $x \in \R$ and some constants $C > 0$, $K \in \mathbb{Z}^+$.
    \item The number of training samples $m$, dimension of input vector $n$ and number of intermediate features $k$ jointly diverge: $m,n,k \to \infty$ with $n/m, n/k \in (0,\infty)$.
    \item $\F := [\f_1, \f_2, \dots, \f_k]^T$ where $\f_i \sim \Normal(0, \frac{1}{n + \theta} \I_n)$. Note that the covariance of $\f_i$ is selected such that $\E_{\x, \f_i}[(\f_i^T\x)^2] = 1$.
    \item The activation function $\sigma: \R \to \R$ is an odd function satisfying $|\sigma(x)| < C (1 + |x|^K)$ for all $x \in \R$ and some constants $C > 0$, $K \in \mathbb{Z}^+$.
\end{enumerate}

\begin{figure*}
    \centering
    \begin{subfigure}[b]{0.35\textwidth}
         \centering
         \includegraphics[width=0.99\linewidth]{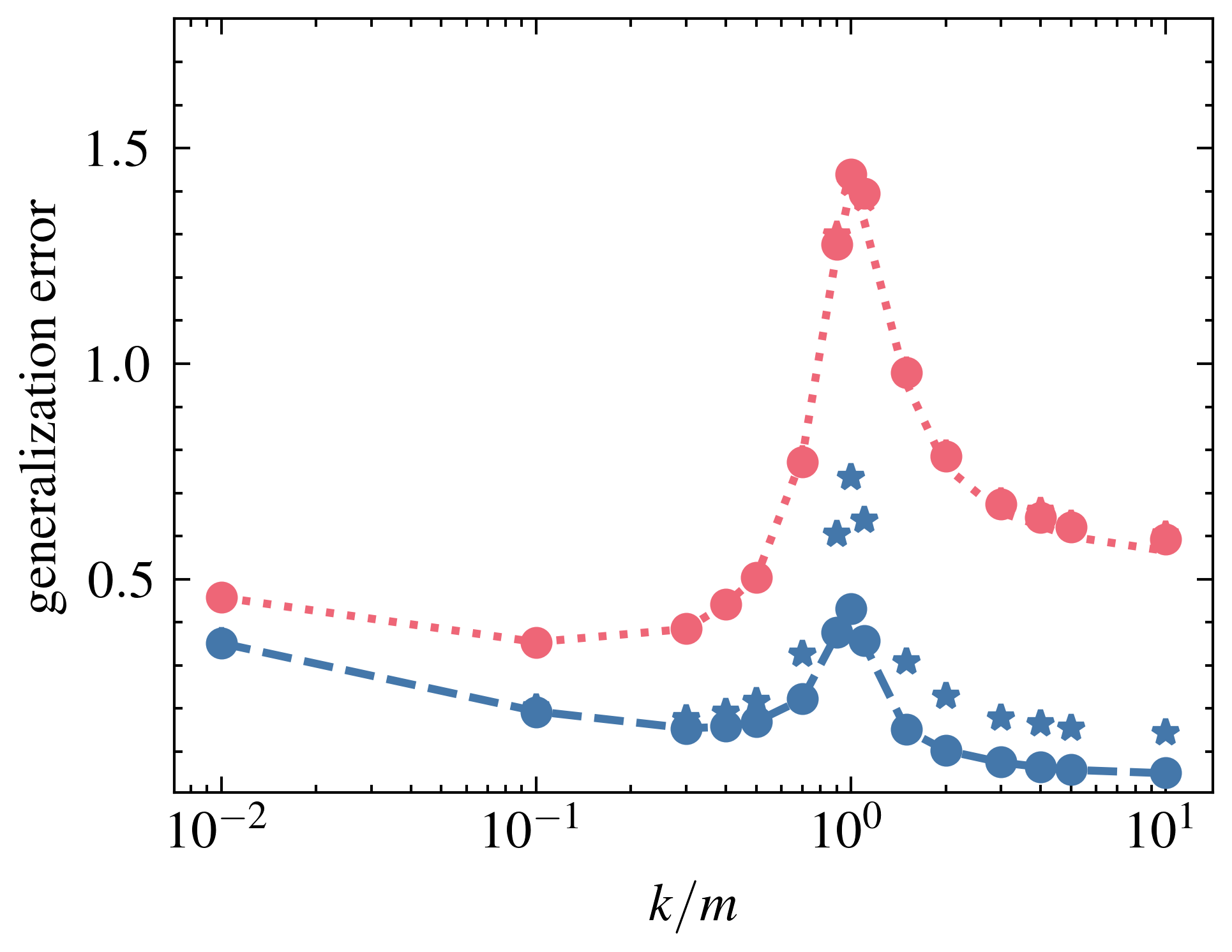}
         \captionsetup{justification=centering}
         \caption{$\sigma_* = \sigma_{ReLU}$}
         \label{figure:relu_polynomial_equivalence}
     \end{subfigure}
     \hspace{2em}
     \begin{subfigure}[b]{0.35\textwidth}
         \centering
         \includegraphics[width=0.99\linewidth]{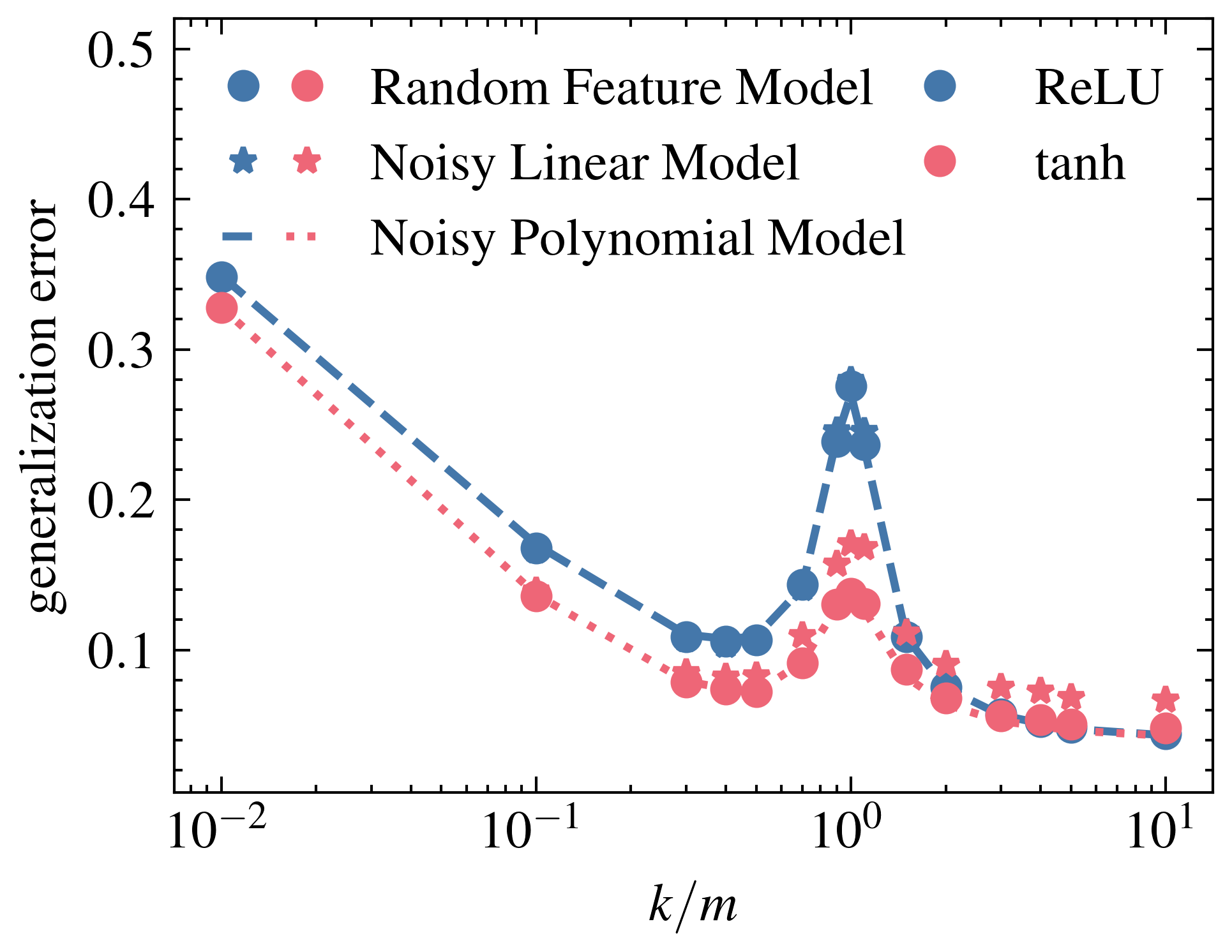}
         \captionsetup{justification=centering}
         \caption{$\sigma_* = \sigma_{tanh}$}
         \label{figure:tanh_polynomial_equivalence}
     \end{subfigure}
     \caption{\textit{The nonlinear side of the phase boundary.} When $\alpha=1$ and $\theta = n^{1/2}$, the RFM crosses the phase boundary and the noisy linear model~\eqref{eq:gaussian_model} no longer suffices. The noisy polynomial model~\eqref{eq:polynomial_model} captures the RFM generalization error accurately. Here $\lambda=10^{-3}$, $n=400$, $m=500$; averages over 50 Monte Carlo runs.}
     \label{fig:polynomial_equivalence}
\end{figure*}

\paragraph{Discussion of Assumptions}
S.1--S.3 define the data model. The key parameter is the input--label
correlation $\alpha = \bgamma^T\bxi$, which, together with the spike
magnitude $\theta$, controls the phase transition we identify. The restriction $\beta < 1/2$ in S.2 is required by our Lindeberg
replacement argument (Appendix~\ref{appendix:equivalence_proof}):
our concentration bounds regarding $\F \x$
(in Appendix~\ref{appendix:admissible_feature_matrices}) fails for $\beta \ge 1/2$. This
regime boundary is inherent to the \emph{unconditional} Gaussian
equivalence framework used here and
in~\cite{hu2022universality}. Recent \emph{conditional} Gaussian
equivalence techniques~\cite{dandi2023learning,
demir2025asymptotic, wen2025does}, which condition on the
spike-aligned component of the features rather than treating them
monolithically, provide a natural path to extending the analysis to
$\beta \geq 1/2$; we expect the phase transition to persist in
this regime, as our experiments already confirm
(Figure~\ref{figure:comparison_theta}). The normalization in S.3 ensures the
teacher input has unit variance, so that label complexity is determined
by $\sigma_*$ alone.  S.4 specifies the proportional asymptotic limit.
In S.5, the $1/(n+\theta)$ scaling guarantees
$\E_{\x,\f_i}[(\f_i^T\x)^2]=1$ regardless of spike magnitude, so
that the Hermite expansion of $\sigma$---and hence the effective
polynomial degree governing the phase transition---is defined with
respect to a standard Gaussian argument. S.6 requires $\sigma$ to be odd with polynomial growth; this is
inherited from~\cite{hu2022universality} and ensures covariance
matching to the linear surrogate~\eqref{eq:gaussian_model}
via~\eqref{eq:covariance_linear}. One may drop the odd assumption in the expanse of replacing the linear
surrogate with a quadratic one, but it does not affect the phase
transition mechanism itself. While ReLU and Softplus
violate the odd-function requirement, all numerical
experiments---including those on CIFAR-10
(Figure~\ref{fig:CIFAR-10})---confirm that the predicted phase
transition persists, suggesting this is a proof artifact rather than a
fundamental limitation.

\section{Main Results}
\label{sec:main_results}

This section develops the theoretical machinery behind the phase transition described in Section~\ref{sec:intro}. We proceed in three steps. First, Section~\ref{sec:universality} establishes a universality theorem under spiked covariance: two activations yield the same RFM performance whenever their low-order moment structures match. Second, Section~\ref{sec:high_order} applies this universality to show that every RFM is performance-equivalent to a \emph{noisy polynomial} model whose degree is controlled by the quantity $\eta$, which couples input--label correlation, spike magnitude, and feature matrix. Third, Sections~\ref{sec:linear} and~\ref{sec:polynomial} use this characterization to delineate the two sides of the phase boundary: the effectively linear regime ($\eta = \mathcal{O}(n^{-1/2})$) and the genuinely nonlinear regime ($\eta$ larger).

\vspace{1em}
\subsection{Universality of Random Features under Spiked Data}
\label{sec:universality}

The classical universality result of~\cite{hu2022universality} shows that, under isotropic data, the activation $\sigma$ can be replaced by a matched Gaussian surrogate~\eqref{eq:gaussian_model} without changing training or generalization performance. This result relies on rotational invariance and does not extend to anisotropic settings. The following theorem provides the required extension: under spiked covariance, two activations are performance-equivalent whenever their first two moment structures---covariance and cross-covariance with the label---are matched.

\begin{theorem}
    Let $\sigma(x)$, $\hat{\sigma}(x)$ be two activation functions satisfying S.6. Suppose that assumptions S.1-S.6 given in Section \ref{sec:assumptions} hold. If
    \begin{align}
           &\left\|\E_{\x} [\sigma(\F\x) \sigma(\F\x)^T]  -  \E_{\x} [\hat{\sigma}(\F\x) \hat{\sigma}(\F\x)^T]\right\| = \mathcal{O}(k^{-\varsigma}), \label{eq:Sigma_x}\\ 
            &\left\|\E_{(\x,y)} \left[\sigma(\F\x) y \right] - \E_{(\x,y)} \left[\hat{\sigma}(\F\x) y \right]\right\| = \mathcal{O}(k^{-\varsigma}) ,\label{eq:Sigma_xy}
    \end{align}
    are satisfied for some $\varsigma >0$, then
    \begin{enumerate}[label=(\roman*)]
        \item the training error $\T_{\sigma}$ of the RFM with activation $\sigma$, and the training error $\T_{\hat{\sigma}}$ of the RFM with activation $\hat{\sigma}$, both converge in probability to the same value $e_{\T} \geq 0$,
        \item the corresponding generalization errors $\G_{\sigma}$ and $\G_{\hat{\sigma}}$ also converge in probability to the same value $e_{\G} \geq 0$ under additional assumptions S.7 and S.8 provided in Appendix \ref{appendix:equivalence_proof}.
    \end{enumerate}
    \begin{proof}
The overall strategy follows the Lindeberg replacement method used
by~\cite{hu2022universality} under isotropy: we define a perturbed
training objective, then bound the expected difference when replacing
$\sigma(\F\x_i)$ with $\hat{\sigma}(\F\x_i)$ one sample at a
time~\cite{lindeberg1922neue,korada2011applications}. Extending this
argument to spiked covariance requires a new \emph{admissible set} of
feature matrices (Appendix~\ref{appendix:admissible_feature_matrices}) that controls
the spike-projected quantities $|\bgamma^T\f_j\,\f_i^T\bgamma|$ and the
spectral norm of $\F(\I_n+\theta\bgamma\bgamma^T)\F^T$, both of which
are trivially bounded under isotropy but require dedicated concentration
arguments under anisotropy. The restriction $\beta < 1/2$ enters here to ensure these
concentration bounds hold. 

The main new difficulty, however, arises not
in the universality statement itself but in \emph{verifying its
conditions}---showing that conditions~\eqref{eq:Sigma_x}
and~\eqref{eq:Sigma_xy} hold for the Hermite surrogate---which is the
content of Theorem~\ref{theorem:high_order_equivalence} and where the
phase transition mechanism emerges (see the proof discussion therein). We
provide the complete proof in
Appendix~\ref{appendix:equivalence_proof}.
    \end{proof}
    \label{theorem:universality}
\end{theorem}

Theorem~\ref{theorem:universality} is more general than a reduction to a single surrogate model: it establishes that \emph{any} two activations with matched moment structure yield identical RFM performance under spiked data. We exploit this generality next by choosing $\hat{\sigma}$ to be a finite-degree Hermite polynomial, which makes the effective nonlinear structure of the RFM explicit.

\vspace{1em}
\subsection{Equivalent Noisy Polynomial Model and the Role of $\eta$}
\label{sec:high_order}

We now identify the surrogate that reveals the phase transition. The key tool is the Hermite expansion. For $i \in \mathbb{Z}^+$, the $i$-th Hermite polynomial is $H_i(x) := (-1)^i e^{x^2/2} \frac{d^i}{dx^i} e^{-x^2/2},$ and any activation $\sigma$ with $\E_{z \sim \Normal(0,1)}[\sigma(z)^2] < \infty$ admits the expansion
\begin{align}
    \sigma(x) = \sum_{i=0}^{\infty} \frac{1}{i!} \mu_i H_i(x), \quad \text{ with } \quad \mu_i := \E_{z \sim \Normal(0,1)}[H_i(z) \sigma(z)]. \label{eq:hermite_expansion}
\end{align}
The following theorem shows that, under spiked covariance, the RFM with activation $\sigma$ is performance-equivalent to an RFM whose activation is a \emph{finite-degree} Hermite polynomial plus noise. Crucially, the degree of this polynomial is governed by a single quantity $\eta$ that encodes the interaction between input--label correlation, spike magnitude, and feature matrix.

\begin{theorem}
    Let $\sigma$ be any activation function satisfying S.6. Define another activation function
    \begin{align}
        \hat{\sigma}_l(x) := \left(\sum_{j=0}^{l-1} \frac{1}{j!} \mu_j H_j(x) \right) + \mu^*_l z \quad \text{with} \quad z \sim \Normal(0,1), \label{eq:equivalent_activation}
    \end{align}
    where $\mu_j := \E_{z \sim \Normal(0,1)}[H_j(z) \sigma(z)]$ and $\mu^*_l := \sqrt{\E_{z \sim \Normal(0,1)}[\sigma(z)^2] - \sum_{j=0}^{l-1} \mu_j^2/ (j!)}$. 
    
    Suppose that assumptions S.1-S.6 given in Section \ref{sec:assumptions} hold. If
    \begin{equation}
        \eta := \max _{1 \leq i \leq k} \left|\frac{(\bxi + \theta \alpha \bgamma)^T \f_i}{\sqrt{1+\theta \alpha^2}}\right| \leq \frac{C}{n^{1/l}}, \quad \text{for some $C > 0$ and some $l \in \mathbb{Z}^+$,} 
        \label{eq:eta}
    \end{equation}
    is satisfied, then
    \begin{enumerate}[label=(\roman*)]
        \item the training error $\T_{\sigma}$ of the RFM with activation $\sigma$, and the training error $\T_{\hat{\sigma}_l}$ of the RFM with activation $\hat{\sigma}_l$, both converge in probability to the same value $e_{\T} \geq 0$,
        \item  the corresponding generalization errors $\G_{\sigma}$ and $\G_{\hat{\sigma}_l}$ also converge in probability to the same value $e_{\G} \geq 0$ under additional assumptions S.7 and S.8 provided in Appendix \ref{appendix:equivalence_proof}.
    \end{enumerate}
    \begin{proof}
        We verify conditions~\eqref{eq:Sigma_x} and~\eqref{eq:Sigma_xy}
        of Theorem~\ref{theorem:universality} for the pair $(\sigma,
        \hat{\sigma}_l)$; the two equivalences require different arguments
        and interact differently with the spike.

        \textit{Covariance equivalence.} In
        Appendix~\ref{appendix:covariance_equivalence}, we work with the
        transformed feature matrix
        $\hat{\F}:=\F(\I_n+(\sqrt{1+\theta}-1)\bgamma\bgamma^T)$, whose
        rows no longer have uniform norms due to the spike. Via a Taylor
        expansion of
        $\E[\sigma(\hat{\f}_i^T\g)\,\sigma(\hat{\f}_j^T\g)]$ with
        remainder bounds that absorb the spike-induced variance in
        $\|\hat{\f}_i\|^2$, we show
        \begin{align}
            \left\|\E_{\x} [\sigma(\F\x) \sigma(\F\x)^T]  -  (\mu_1^2 \F(\I_n + \theta \bgamma \bgamma^T)\F^T + \mu_*^2 \I_k) \right\| = \mathcal{O}(k^{-\varsigma}), 
            \label{eq:covariance_linear}
        \end{align}
        for some $\varsigma > 0$ and any $\sigma$ satisfying S.6, with
        $\mu_1 = \E[z \sigma(z)]$ and $\mu_* = \sqrt{\E[\sigma(z)^2] -
        \mu_1^2}$ for $z \sim \Normal(0,1)$. Since $\hat{\sigma}_l$
        preserves $\mu_1$ and $\mu_*$ by construction,
        \eqref{eq:covariance_linear} holds for both $\sigma$ and
        $\hat{\sigma}_l$ with the same right-hand side. The triangle
        inequality then gives
        \begin{align}
            \left\|\E_{\x} [\sigma(\F\x) \sigma(\F\x)^T]  -  \E_{\x} [\hat{\sigma}_l(\F\x) \hat{\sigma}_l(\F\x)^T]\right\| = \mathcal{O}(k^{-\varsigma}). \label{eq:Sigma_x_hermite}
        \end{align}
        Crucially, the covariance equivalence holds for \emph{all}
        activations satisfying S.6, with no dependence on $\eta$; the
        spike affects only the remainder bounds, not the structure of the
        result.

        \textit{Cross-covariance equivalence.} This is where the phase
        transition originates. In
        Appendix~\ref{appendix:cross_covariance_equivalence}, we
        decompose $\E[\sigma(\f_i^T\x)\,y]$ using the Hermite expansions
        of $\sigma$ and $\sigma_*$. Under isotropy, each feature
        $\f_i^T\x$ and the label projection $\bxi^T\x$ are nearly
        independent---their correlation is
        $\mathcal{O}(n^{-1/2})$---so only the first Hermite coefficient
        contributes, enforcing effective linearity. Under spiked
        covariance, the correlation $\eta_i =
        (\f_i^T\bxi+\theta\alpha\,\f_i^T\bgamma)/\sqrt{1+\theta\alpha^2}$
        can be substantially larger due to the $\theta\alpha$
        amplification term, and the Hermite decomposition retains
        contributions up to order $l$ whenever $\eta^l \ge \Tilde C
        n^{-1}$ for some $\Tilde C >0$. Bounding the remainder of this expansion under
        the condition $\eta = \mathcal{O}(n^{-1/l})$ yields
        \begin{equation}
            \left\|\E_{(\x,y)} \left[\sigma(\F\x) y \right] - \E_{(\x,y)} \left[\hat{\sigma}_l(\F\x) y \right]\right\| = \mathcal{O}(k^{-\varsigma}). \label{eq:Sigma_xy_hermite}
        \end{equation}
        Applying Theorem~\ref{theorem:universality} with
        \eqref{eq:Sigma_x_hermite} and \eqref{eq:Sigma_xy_hermite}
        completes the proof.
    \end{proof}
    \label{theorem:high_order_equivalence}
\end{theorem}

\paragraph{Interpretation: $\eta$ as the phase transition parameter}
Theorem~\ref{theorem:high_order_equivalence} shows that the effective polynomial degree of the RFM is not intrinsic to the activation but is \emph{selected by the data} through $\eta$. The quantity $\eta$ is directly related to the input--label correlation: it depends on $(\bxi + \theta\alpha\bgamma)/\sqrt{1+\theta\alpha^2}$, which appears as the leading term in the decomposition
\begin{align}
    \E[(\x - \E[\x])  (y - \E[y])] = \Tilde{\mu}_1 \frac{\bxi + \theta \alpha \bgamma}{\sqrt{1+ \theta \alpha^2}} + \sum_{j>1} \Tilde{\mu}_j \E_{\x}[\x H_j(\bxi^T \x)], \label{eq:input_label_correlation}
\end{align} 
where $\Tilde{\mu}_j := \E_{z \sim \Normal(0,1)}[\sigma_*(z) H_j(z)]$. A larger value of $\|\bxi + \theta\alpha\bgamma\|/\sqrt{1+\theta\alpha^2}$ implies a larger $\eta$ for any $\F$ sampled independently of $\bgamma$ and $\bxi$, and hence a higher effective polynomial degree. In particular, strong alignment ($\alpha$ large) combined with large spike magnitude ($\theta$ large) pushes $\eta$ above the linear threshold, activating higher-order Hermite terms. This is the mechanism behind the phase transition.

\subsection{The Linear Side of the Phase Boundary}
\label{sec:linear}

Setting $l=2$ in Theorem~\ref{theorem:high_order_equivalence} reduces the equivalent activation to $\hat{\sigma}_1(x) = \mu_0 + \mu_1 x + \mu_2^* z$, recovering the classical noisy linear surrogate~\eqref{eq:gaussian_model}. The following corollary gives the precise condition under which the RFM remains on the linear side of the phase boundary.

\begin{corollary}
\label{cor:linear}
The noisy linear model \eqref{eq:gaussian_model} performs equivalent to the RFM with activation $\sigma$ in terms of training and generalization errors if $\eta = \mathcal{O}(n^{-1/2})$ in \eqref{eq:eta}.
\end{corollary}

\begin{remark}
\label{remark:boundary_sufficiency}
Corollary~\ref{cor:linear} provides a \emph{sufficient} condition for
effective linearity. The boundary $\eta = \mathcal{O}(n^{-1/2})$ in
Figure~\ref{fig:phase_diagram} is therefore an upper bound on the
linear regime: below it, the RFM is provably equivalent to the noisy
linear model; above it, higher-order terms are \emph{permitted} by
the theory and are consistently observed in all our experiments
(Figures~1--4). 
\end{remark}

Figure~\ref{fig:phase_diagram} displays the phase diagram. The red curve traces the boundary $\eta = \mathcal{O}(n^{-1/2})$ in the $(\alpha,\theta)$~plane: below this curve, the RFM and the noisy linear model have indistinguishable generalization errors; above it, a gap opens and grows with $\alpha$ and $\theta$. Since $\f_i$ is sampled independently of $\bgamma$ and $\bxi$, both $|\bgamma^T\f_i|$ and $|\bxi^T\f_i|$ scale as $\mathcal{O}(n^{-1/2})$. The condition $\eta = \mathcal{O}(n^{-1/2})$ therefore reduces to $|\theta\alpha/\sqrt{1+\theta\alpha^2}| = \mathcal{O}(1)$.

As a concrete example, consider the misaligned case $\alpha = \mathcal{O}(n^{-1/2})$, which arises with high probability when $\bgamma$ and $\bxi$ are drawn independently from $\Normal(0,\I_n)$ and normalized. Then $\eta = \mathcal{O}(n^{-1/2})$ holds for all $\theta \asymp n^\beta$ with $\beta \in [0,1/2)$, placing the RFM firmly in the linear regime. Figure~\ref{fig:linear_equivalence} confirms this: the generalization errors of the RFM and the noisy linear model coincide across activation functions.

\subsection{The Nonlinear Side: Higher-Order Polynomial Models}
\label{sec:polynomial}

When $\eta$ exceeds the linear threshold of Corollary~\ref{cor:linear}, the RFM crosses the phase boundary and the noisy linear surrogate no longer captures its performance. Theorem~\ref{theorem:high_order_equivalence} provides the correct description on this side: the RFM is equivalent to the noisy polynomial model
\begin{equation}
    \w^T \hat{\sigma}_l(\F \x), \label{eq:polynomial_model}
\end{equation}
for $l > 2$, where $\hat{\sigma}_l$ is defined in~\eqref{eq:equivalent_activation}. The higher-order Hermite terms that were invisible in the linear regime now persist and actively reduce generalization error beyond what any linear predictor can achieve.

Which polynomial orders contribute depends not only on $\eta$ but also on the interplay between the Hermite spectra of the activation $\sigma$ and the teacher $\sigma_*$, as the following remark makes precise.

\begin{remark}
\label{remark:zero_coeff_reduction}
When $\eta = \mathcal{O}(n^{-1/4})$, the equivalent activation is a third-degree polynomial. Let $\mu_0, \mu_1,\mu_2, \mu_3$ and $\Tilde{\mu}_0, \Tilde{\mu}_1,\Tilde{\mu}_2,\Tilde{\mu}_3$ denote the first four Hermite coefficients~\eqref{eq:hermite_expansion} of $\sigma$ and $\sigma_*$, respectively. Then $\E[\hat{\sigma}_l(\f_i^T\x) y] = \sum_{j=0}^{3} \frac{1}{j!} \mu_j \Tilde{\mu}_j \eta_i^j$ by~\eqref{eq:Sigma_xy_decomposition}. The polynomial model~\eqref{eq:polynomial_model} therefore reduces to second degree if $\mu_3\Tilde{\mu}_3 = 0$, and collapses all the way to the noisy linear model~\eqref{eq:gaussian_model} if additionally $\mu_2\Tilde{\mu}_2 = 0$. Thus, even above the phase boundary, the effective degree can be lower than the maximum permitted by $\eta$ when the Hermite spectra of $\sigma$ and $\sigma_*$ are ``mismatched'' at certain orders.
\end{remark}

Figure~\ref{fig:polynomial_equivalence} illustrates Remark~\ref{remark:zero_coeff_reduction} in the aligned case ($\alpha = 1$, $\theta = n^{1/2}$), where the RFM is firmly on the nonlinear side of the phase boundary. In Figure~\ref{figure:relu_polynomial_equivalence} ($\sigma_* = \sigma_{ReLU}$), the noisy linear model suffices for tanh activation but not for ReLU, because tanh has $\mu_2 = 0$ while ReLU has $\mu_2 \neq 0$. Conversely, in Figure~\ref{figure:tanh_polynomial_equivalence} ($\sigma_* = \sigma_{tanh}$), the situation reverses: tanh activation requires the polynomial surrogate ($\mu_3 \neq 0$) while ReLU does not ($\mu_3 = 0$). In both cases, the noisy polynomial model~\eqref{eq:polynomial_model} accurately tracks the RFM. These results highlight that the effective degree on the nonlinear side of the phase boundary is jointly determined by the Hermite coefficients of $\sigma$ and $\sigma_*$---the label-generation process is as important as the activation in determining which nonlinear terms contribute.

\begin{figure*}
    \centering
    \begin{subfigure}[b]{0.325\textwidth}
         \centering
         \includegraphics[width=0.99\linewidth]{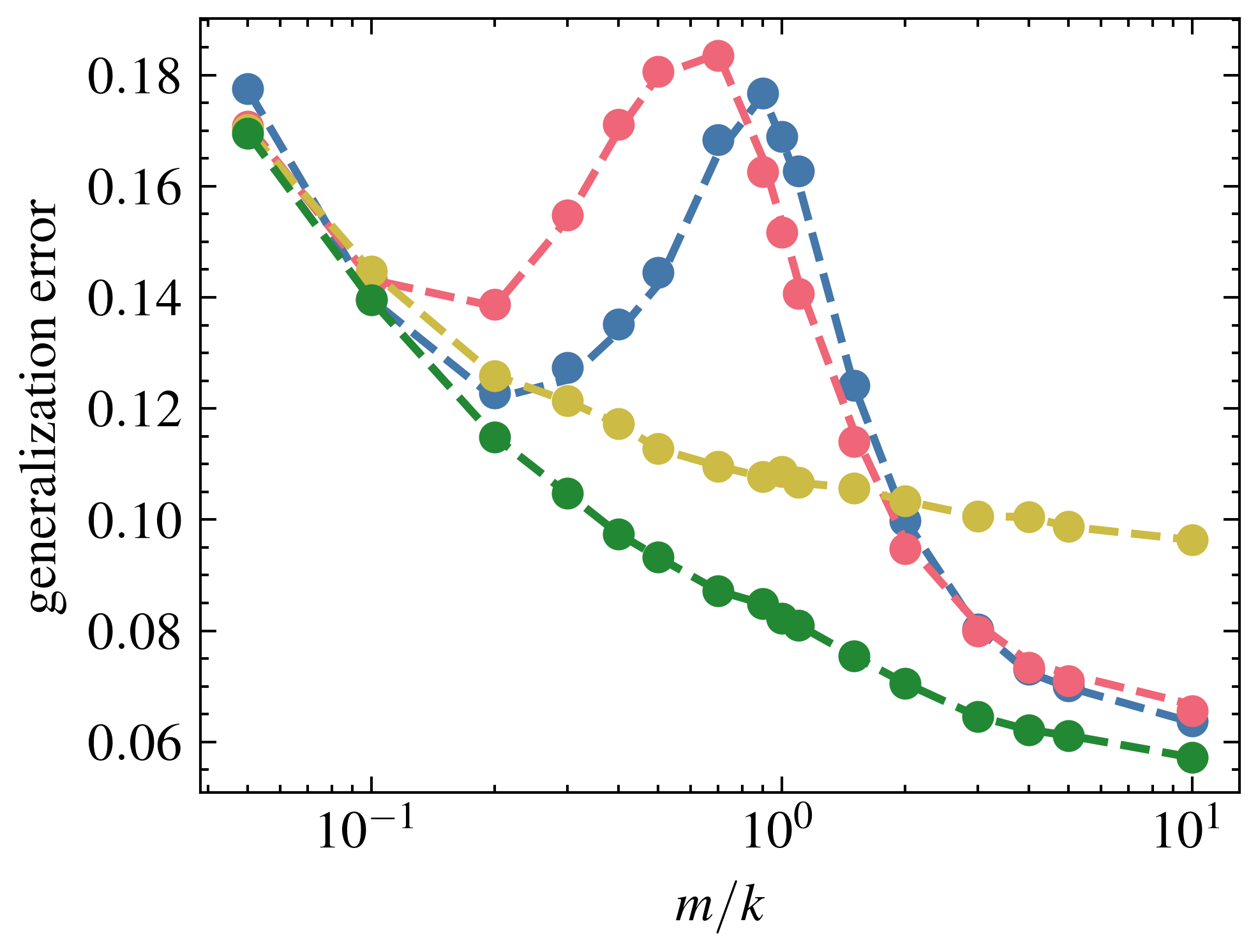}
         \captionsetup{justification=centering}
         \caption{impact of number of samples\\($\alpha = 1$ and $\theta=n^{1/2}$)}
         \label{figure:comparison_samples}
     \end{subfigure}
     \hfill
     \begin{subfigure}[b]{0.325\textwidth}
         \centering
         \includegraphics[width=0.99\linewidth]{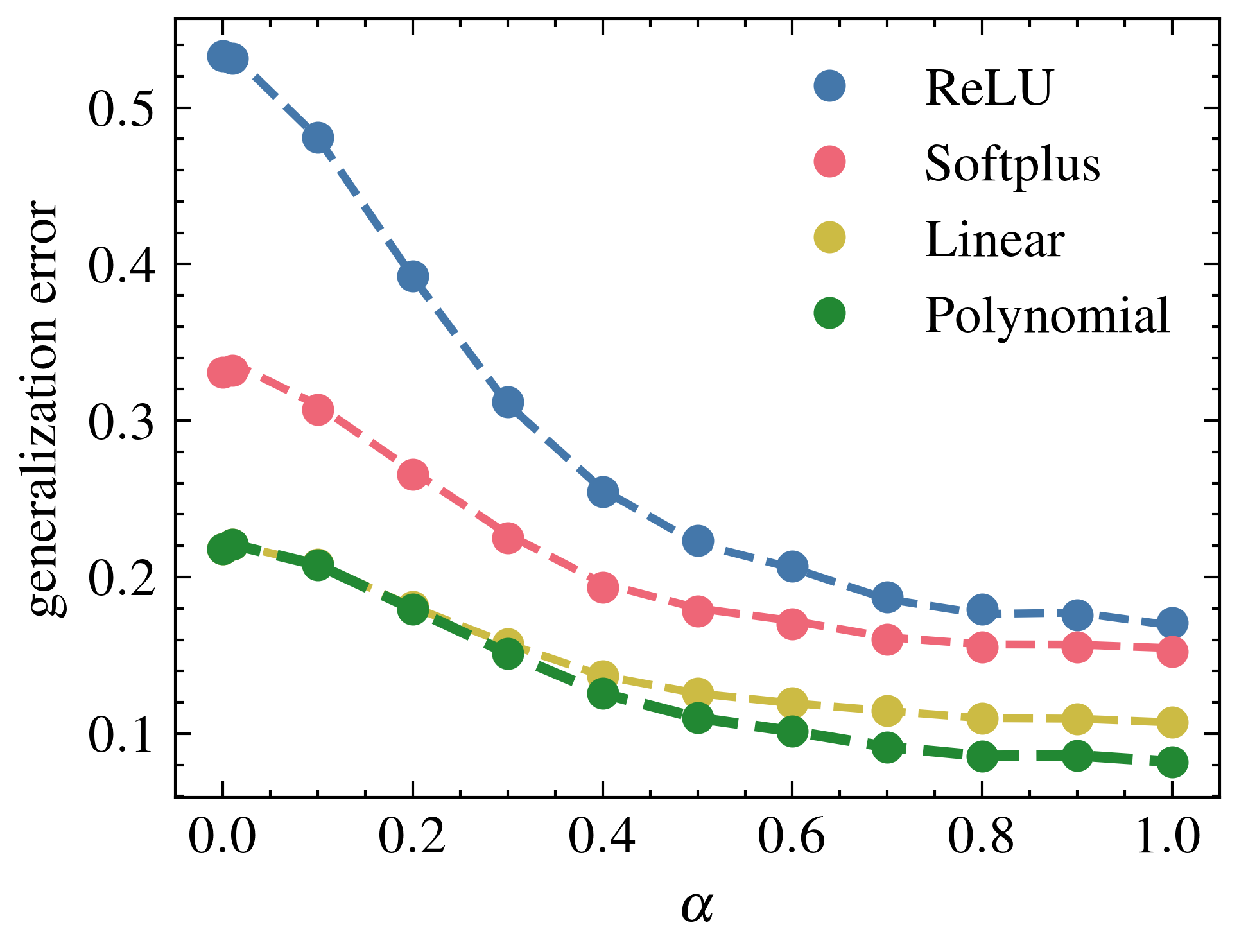}
         \captionsetup{justification=centering}
         \caption{impact of alignment\\($m=500$ and $\theta=n
         ^{1/2}$)}
         \label{figure:comparison_alignment}
     \end{subfigure}
     \hfill
     \begin{subfigure}[b]{0.325\textwidth}
         \centering
         \includegraphics[width=0.99\linewidth]{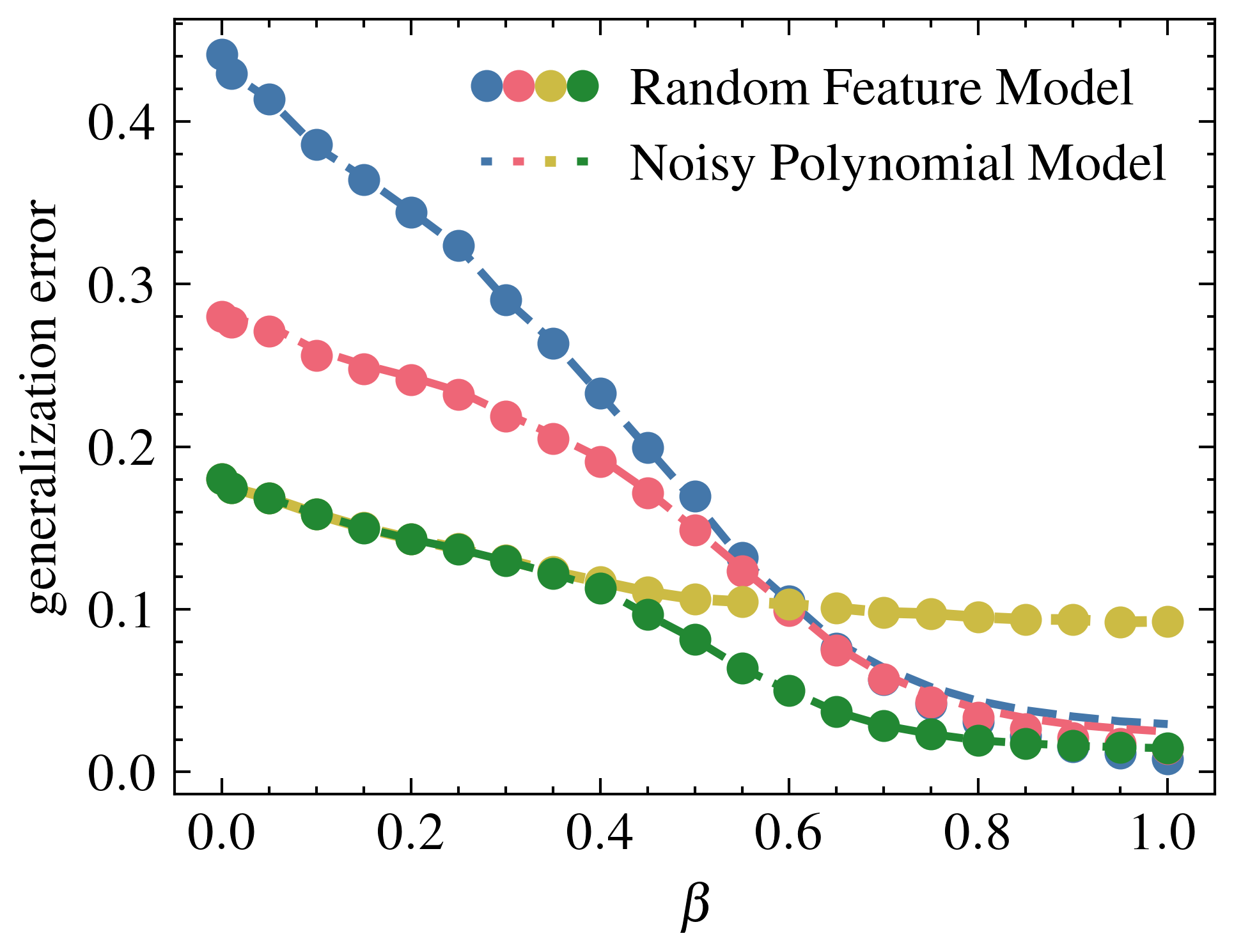}
         \captionsetup{justification=centering}
         \caption{impact of $\theta \asymp n^\beta$\\($m=500$ and $\alpha = 1$)}
         \label{figure:comparison_theta}
     \end{subfigure}
     \caption{\textit{Comparison of activation functions} - generalization errors of the RFM with different nonlinearities (linear, polynomial, Softplus, ReLU) with respect to number of samples (on the left), alignment (on the center), and spike magnitude $\theta \asymp n^\beta$ (on the right). Here, $n=400$, $k=500$, $\lambda=10^{-2}$, and $\sigma_*=\sigma_{ReLU}$. We limit the degree of the equivalent polynomial model \eqref{eq:equivalent_activation} to a maximum of $l=4$ for numerical stability. We plot the average of 50 Monte Carlo runs.}
     \label{fig:comparison}
\end{figure*}

\section{Crossing the Phase Boundary: Nonlinear Gains in Practice}
\label{sec:crossing}

The phase transition identified in Section~\ref{sec:main_results} predicts that the RFM achieves genuinely nonlinear gains when the input--label correlation and spike magnitude push $\eta$ above the linear threshold. We now verify this prediction numerically and quantify the gain across activation functions.

\paragraph{Baselines}
To isolate the effect of nonlinearity, we compare against two oracle baselines whose coefficients are tuned numerically to minimize generalization error. The first is an optimal linear activation,
\begin{align}
\sigma_{linear}(x) = a_0 + a_1 x,
\label{eq:optimal_linear}
\end{align}
with $a_0, a_1 \in \R$. This represents the best possible performance achievable by any linear predictor in the RFM framework. The second is an optimal third-degree polynomial activation,
\begin{align}
\sigma_{polynomial}(x) = b_0 + b_1 x + \tfrac{1}{2} b_2 (x^2 - 1) + \tfrac{1}{6} b_3 (x^3 - 3x) + b_4 z,
\label{eq:optimal_polynomial}
\end{align}
with $z \sim \Normal(0,1)$ and $b_0,\dots,b_4 \in \R$ tuned analogously. This baseline captures the maximum performance permitted by Theorem~\ref{theorem:high_order_equivalence} under $\eta = \mathcal{O}(n^{-1/4})$, which holds with high probability when $\beta < 1/2$.

\paragraph{Impact of sample size}
Figure~\ref{figure:comparison_samples} compares generalization errors as a function of $m/k$ in the aligned case ($\alpha = 1$, $\theta = n^{1/2}$)---firmly on the nonlinear side of the phase boundary. The polynomial RFM consistently achieves the lowest generalization error across the full range of $m/k$, confirming that higher-order terms provide a genuine advantage in this regime. ReLU and Softplus exhibit double descent~\cite{Belkin19,Geiger_2020}, causing the optimized linear activation to outperform them in the mid-range of $m/k$. However, the polynomial baseline avoids double descent entirely (its coefficients are optimized) and dominates throughout. These patterns are consistent with findings under isotropic data~\cite{wang2022optimal,demir2023optimal}, but here the nonlinear gain itself is a consequence of being above the phase boundary.

\begin{figure*}
    \centering
    \begin{subfigure}[b]{0.35\textwidth}
         \centering
         \includegraphics[width=0.99\linewidth]{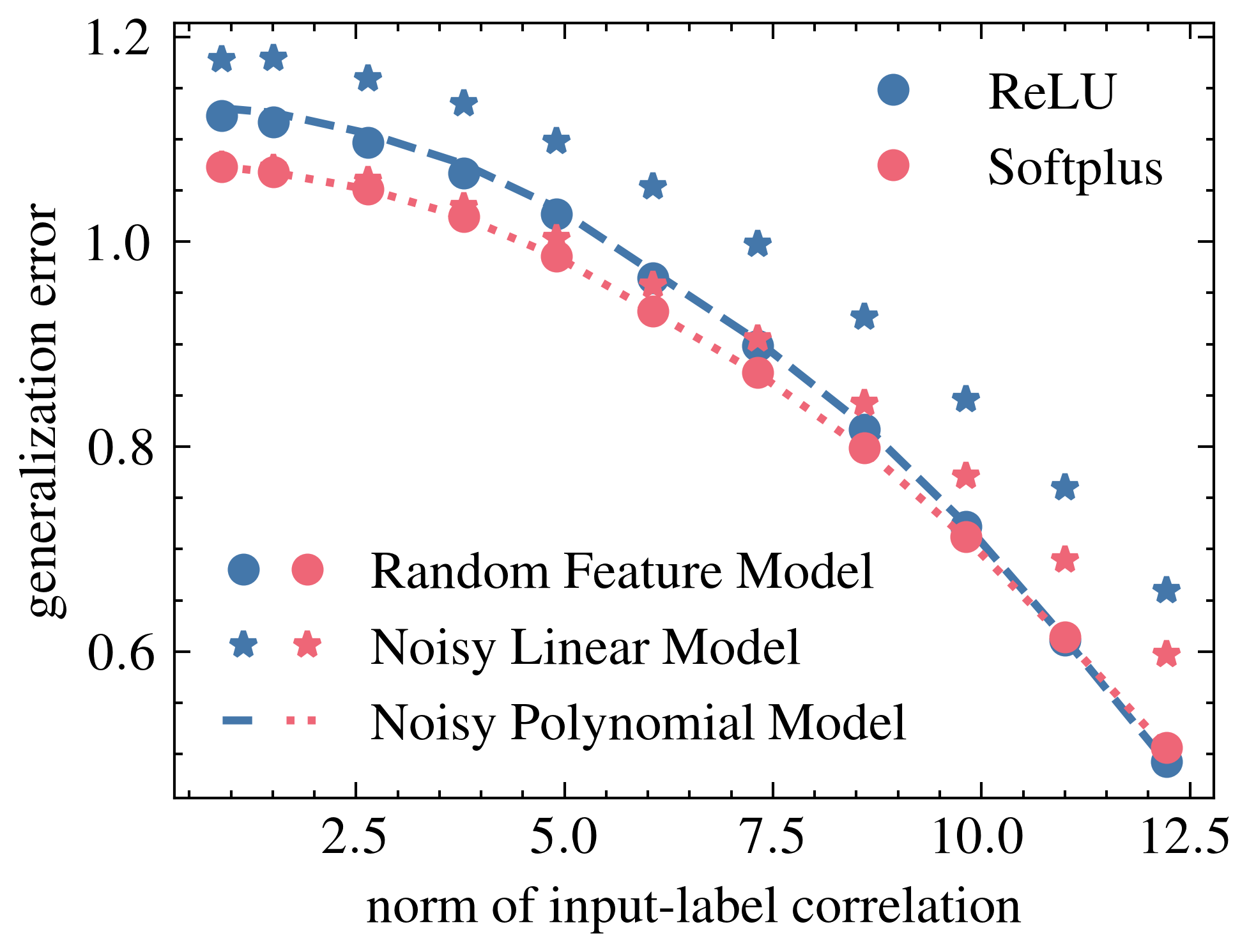}
         \captionsetup{justification=centering}
         \caption{Original inputs}
     \end{subfigure}
     \hspace{2em}
     \begin{subfigure}[b]{0.35\textwidth}
         \centering
         \includegraphics[width=0.99\linewidth]{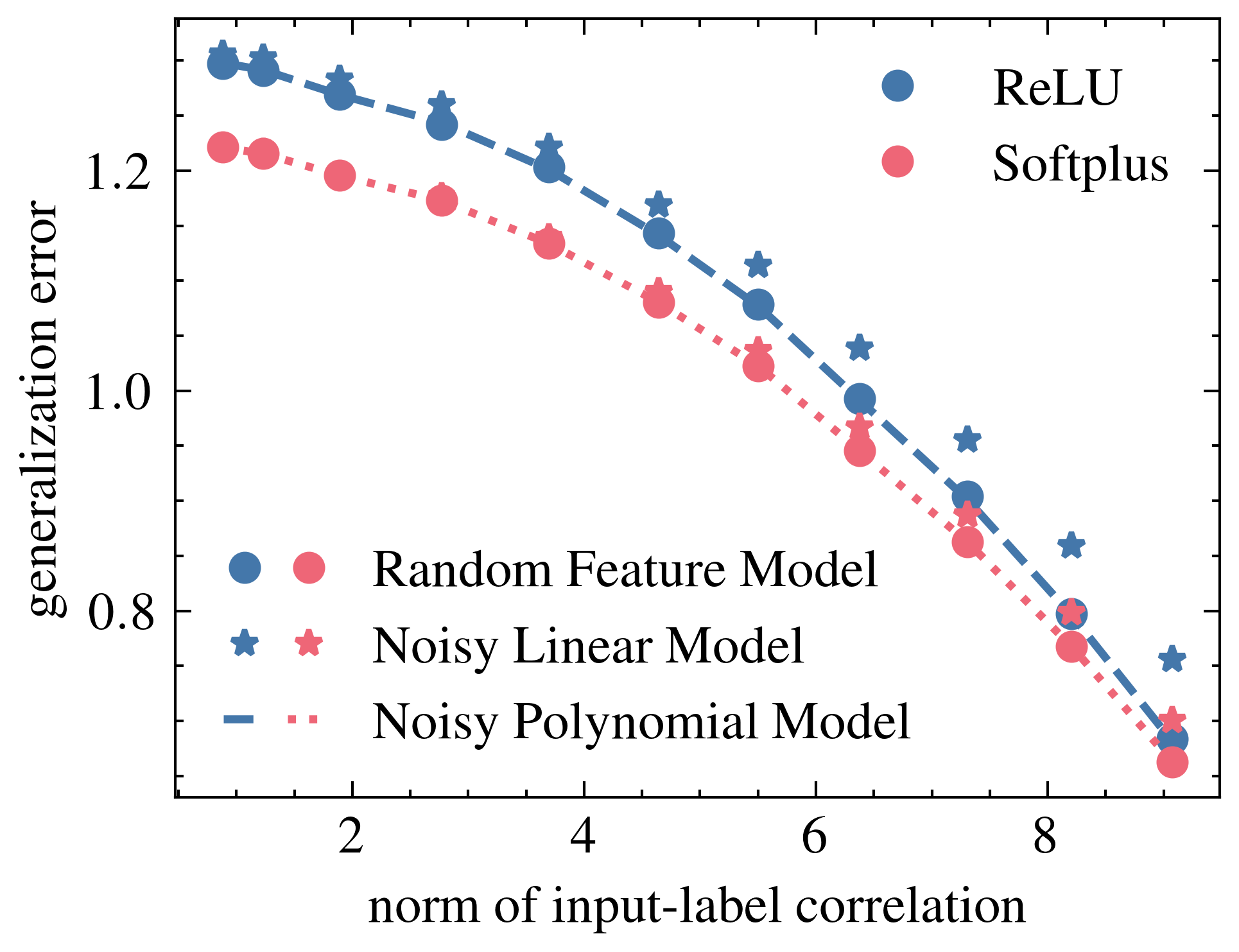}
         \captionsetup{justification=centering}
         \caption{Inputs with Gaussian noise}
     \end{subfigure}
     \caption{\textit{The phase transition on real data: CIFAR-10 classification (airplane vs.\ automobile).} Input--label correlation is controlled by interpolating between true and random labels; the highest correlation norm uses true labels, the lowest uses random labels. As correlation increases, the RFM separates from the linear model, consistent with the predicted phase transition. Adding Gaussian noise (right) isolates the correlation effect. Here $l = 5$, $\lambda = 10^{-1}$, $k=n=3072$, $m=4000$; averages over 50 Monte Carlo runs. Details in Supplementary~\ref{supplement:CIFAR-10}.}
     \label{fig:CIFAR-10}
\end{figure*}

\paragraph{Impact of alignment}
Figure~\ref{figure:comparison_alignment} varies $\alpha$ while holding $\theta = n^{1/2}$ fixed, sweeping across the phase boundary. For $\alpha \leq 0.3$ the RFM is in the linear regime: the polynomial and linear baselines yield nearly identical errors, confirming that higher-order terms contribute negligibly. A gradual separation emerges for $0.3 < \alpha \leq 0.6$, and for $\alpha > 0.6$ the polynomial RFM distinctly outperforms the linear one. This progression directly traces the transition predicted by Theorem~\ref{theorem:high_order_equivalence}: as $\alpha$ increases, $\eta$ grows, higher Hermite orders activate, and the nonlinear gain materializes.

\paragraph{Impact of spike magnitude}
Figure~\ref{figure:comparison_theta} varies the spike exponent $\beta$ while holding $\alpha = 1$ fixed. Generalization errors for all activations decrease as the spike magnitude grows. For $\beta < 0.4$ the linear and polynomial baselines are indistinguishable---$\eta$ remains near the linear threshold despite full alignment. As $\beta$ increases past $0.4$, the polynomial RFM separates from the linear one, confirming that the phase transition requires both sufficient alignment \emph{and} sufficient spike magnitude. Notably, ReLU and Softplus continue to track their equivalent noisy polynomial models even beyond $\beta = 0.5$, suggesting the phase boundary extends beyond the range covered by our proofs.

\section{Real-World Validation}
\label{sec:real_world}

To test whether the predicted phase transition manifests beyond the spiked-covariance model, we study binary classification (airplane vs.\ automobile) on CIFAR-10~\cite{krizhevsky2009learning}. Figure~\ref{fig:CIFAR-10}a shows results on the original inputs. The noisy polynomial model tracks the RFM across all correlation levels, while the gap between the RFM and the noisy linear model grows with correlation strength---mirroring the phase transition on synthetic data. Furthermore, to isolate the role of input--label correlation from other structural properties of the covariance, we add standard Gaussian noise to the inputs so that the linear model matches the RFM at weak correlation. Figure~\ref{fig:CIFAR-10}b confirms that, even in this controlled setting, the generalization errors of the RFM and the linear model progressively separate as correlation increases. This demonstrates that the correlation-driven transition identified
by our theory is not an artifact of the spiked model but reflects a
mechanism that operates on real data. More broadly, many real-world
datasets exhibit low intrinsic
dimensionality~\cite{facco2017estimating,spigler2020asymptotic},
so that their covariance is well-approximated by a
low-rank-plus-identity structure. Hence, the single-spike model
captures the dominant mode of anisotropy, and the correlation
between this mode and the label direction plays precisely the role
of $\alpha$ in our theory.

\section{Conclusion}
\label{sec:conclusion}
We have identified a correlation-driven transition in the effective
nonlinearity of random feature models under spiked covariance. The transition is governed by a single quantity $\eta$ that couples input--label correlation, spike magnitude, and feature matrix: when $\eta = \mathcal{O}(n^{-1/2})$ the RFM is performance-equivalent to a noisy linear predictor; when $\eta$ is larger, higher-order Hermite terms persist and the RFM achieves genuinely nonlinear gains. This characterization rests on two technical contributions---a universality theorem that extends beyond isotropic data, and an equivalent noisy polynomial surrogate whose effective degree is selected by the data through $\eta$---and is corroborated by numerical simulations and real-data experiments on CIFAR-10.

\section*{Acknowledgments}
This work is supported in part by the TÜBİTAK-ARDEB 1001 program under project 124E063. We extend our gratitude to TÜBİTAK for its support. S.D. is supported by an AI Fellowship provided by Koç University \& İş Bank Artificial Intelligence (KUIS AI) Research Center and a PhD Scholarship (BİDEB 2211) from TÜBİTAK.

\appendix

\section{Proof of Universality of Random Features for the Spiked Data}
\label{appendix:equivalence_proof}
Here, we prove Theorem~\ref{theorem:universality} by extending the
Lindeberg replacement of~\cite{hu2022universality} to the
spiked covariance. For a given set of training samples $\{(\mathbf{x}_i, y_i)\}_{i=1}^m$, let $\bR := [\br_1, \br_2, \dots, \br_m]^T$ where $\br_i := \sigma(\F \x_i)$ for all $i \in \{1,2,\dots,m\}$. For a given $\bR$, we introduce the following perturbed optimization:
\begin{align}
    \Phi_{\bR}(\tau_1, \tau_2) := \inf_{\w \in \R^k} \left\{ \sum_{i=1}^m \left(\w^T \br_i - y_{i}\right)^2 + Q(\w) \right\}, \label{eq:perturbed_objective}
\end{align}
where $Q(\w) := m\lambda||\w||_2^2 + \tau_1 k \w^T \Sigma_x \w + \tau_2 k \w^T \Sigma_{xy}$ with $\Sigma_x := \E_{(\x,y)} [\sigma(\F \x) \sigma(\F \x)^T]$ and $\Sigma_{xy} := \E_{(\x,y)} \left[\sigma(\F\x) y \right]$. Here, $Q(\w)$ is a function of $\tau_1, \tau_2$ (in addition to $\w$) but we do not include them in the notation for simplicity.  Note that $\Phi_{\bR}(0, 0)/m$ is equal to the training error ($\T_\sigma$) of the original problem \eqref{eq:train}. The additional terms in $Q(\w)$ are related to the generalization error and we discuss them later. Since we are interested in two different functions $\sigma$ and $\hat{\sigma}$, we define $\A := [\a_1, \a_2, \dots, \a_m]^T$ and $\B := [\b_1, \b_2, \dots, \b_m]^T$ where $\a_i :=\sigma(\F \x_i)$ and $\b_i := \hat{\sigma}(\F \x_i)$ for all $i \in \{1,2,\dots,m\}$. Then, the training error $\T_{\sigma}$ for the RFM with activation $\sigma$, and the training error $\T_{\hat{\sigma}}$ for the case of activation $\hat{\sigma}$ converge in probability to the same value $e_{\T} \geq 0$ if the following holds
\begin{align}
    \frac{\Phi_{\A}(0, 0) }{m} \stackrel{\Prob}{\to} e_{\T} \quad \text{and} \quad \frac{\Phi_{\B}(0, 0) }{m} \stackrel{\Prob}{\to} e_{\T}.
\end{align}
Since $\Phi_{\bR}(0, 0)/m = (k/m) (\Phi_{\bR}(0, 0)/k)$ and $(k/m) \in (0,\infty)$ by assumption S.4, we focus on $\Phi_{\bR}(0, 0)/k$ for the rest of the proof in order to write the bounds in terms of $k$ only.   

Now, we describe how we handle the generalization error. We introduce $\tau_1 k \w^T \Sigma_x \w + \tau_2 k \w^T \Sigma_{xy}$ into $Q(\w)$ to be used in the calculation of the generalization error. Before describing the relationship between the additional terms and the generalization error, we need the following two additional assumptions for the study of generalization error.

\paragraph{Additional Suppositions for Generalization Error} 
\begin{enumerate}[label=S.\arabic* ---]
    \setcounter{enumi}{6}
    \item There exists $\tau_* = \mathcal{O}(1/\sqrt{k})$ and $s \in (0,\infty)$ such that $Q(\w)/m$ is $s$-strongly convex for all $\tau_1, \tau_2 \in (-\tau_*, \tau_*)$.
    \item There exists a limiting function $q^*(\tau_1, \tau_2)$ such that $\Phi_{\bR}(\tau_1, \tau_2)/k \stackrel{\Prob}{\to} q^*(\tau_1, \tau_2)$ for all $\tau_1, \tau_2 \in (-\tau_*, \tau_*)$. Furthermore, the partial derivatives of $q^*(\tau_1, \tau_2)$ exist at $\tau_1 = \tau_2 = 0$, denoted as  $\frac{\partial}{\partial \tau_1} q^*(0,0) = \rho^*$ and $\frac{\partial}{\partial \tau_2} q^*(0,0) = \pi^*$.
\end{enumerate}

Assumption S.7 is required to keep the perturbed objective \eqref{eq:perturbed_objective} convex and bound the norm of optimal $\w$ of \eqref{eq:perturbed_objective}. Based on the assumption S.8, the generalization error of the RFM with $\sigma$ can be specified as follows $\G_\sigma \stackrel{\Prob}{\to} \rho^* -2 \pi^* + \E_y[y^2]$. Furthermore, the generalization error $\G_{\sigma}$ for the RFM with activation $\sigma$ and the generalization error $\G_{\hat{\sigma}}$ for the case of activation $\hat{\sigma}$ converge in probability to the same value $e_{\G} \geq 0$ if the following holds
\begin{align}
    \frac{\Phi_{\A}(\tau_1, \tau_2) }{k} \stackrel{\Prob}{\to} \phi \quad \text{and} \quad \frac{\Phi_{\B}(\tau_1, \tau_2) }{k} \stackrel{\Prob}{\to} \phi, \label{eq:probabilistic_convergence}
\end{align}
for any $\phi \in \R$ and any $\tau_1, \tau_2 \in (-\tau_*, \tau_*)$, a proof which is provided in Supplementary Lemma \ref{supplement_lemma:perturbed_equivalence}. The rest of this appendix shows that \eqref{eq:probabilistic_convergence} holds, which then implies the statement of Theorem \ref{theorem:universality}. To show the convergence in probability \eqref{eq:probabilistic_convergence}, we take advantage of a test function $\varphi(x)$. Specifically, $\eqref{eq:probabilistic_convergence}$ holds if
\begin{align}
    \left| \E \varphi \left( \frac{1}{k} \Phi_\A \right) - \E \varphi \left( \frac{1}{k} \Phi_\B \right) \right| \leq \max \left\{||\varphi||_\infty, ||\varphi^\prime||_\infty, ||\varphi^{\prime\prime}||_\infty \right\} \kappa_k,
    \label{eq:test_difference}
\end{align}
for some $\kappa_k = o(1)$ and every bounded test function $\varphi(x)$ which has bounded first two derivatives (for a proof, we refer to the Supplementary Lemma \ref{supplement_lemma:test_function_equivalence}). From now on, we use $\Phi_{\bR}$ to denote $\Phi_{\bR}(\tau_1, \tau_2)$. We can continue as follows
\begin{align}
\biggl|\mathbb{E} \varphi\left(\frac{1}{k} \Phi_{\boldsymbol{A}}\right)-&\mathbb{E} \varphi\left(\frac{1}{k} \Phi_{\boldsymbol{B}}\right)\biggr| 
\leq \mathbb{E}\left|\mathbb{E}_{\backslash \boldsymbol{F}}\left[\varphi\left(\frac{1}{k} \Phi_{\boldsymbol{A}}\right)\right]-\mathbb{E}_{\backslash \boldsymbol{F}}\left[\varphi\left(\frac{1}{k} \Phi_{\boldsymbol{B}}\right)\right]\right|, \nonumber\\
&\qquad\quad\quad=  \mathbb{E}\left[\left|\mathbb{E}_{\backslash \boldsymbol{F}}\left[\varphi\left(\frac{1}{k} \Phi_{\boldsymbol{A}}\right)\right]-\mathbb{E}_{\backslash \boldsymbol{F}}\left[\varphi\left(\frac{1}{k} \Phi_{\boldsymbol{B}}\right)\right]\right|\left(\mathds{1}_{\mathcal{A}}(\boldsymbol{F})+\mathds{1}_{\mathcal{A}^c}(\boldsymbol{F})\right) \right],\nonumber\\
&\qquad\quad\quad\leq \sup _{\boldsymbol{F} \in \mathcal{A}}\left|\mathbb{E}_{\backslash \boldsymbol{F}}\left[\varphi\left(\frac{1}{k} \Phi_{\boldsymbol{A}}\right)\right]-\mathbb{E}_{\backslash \boldsymbol{F}}\left[\varphi\left(\frac{1}{k} \Phi_{\boldsymbol{B}}\right)\right]\right|+2\|\varphi\|_{\infty} \mathbb{P}\left(\mathcal{A}^c\right),
\end{align}
where $\mathbb{E}_{\backslash \boldsymbol{F}}[\cdot]$ denotes the conditional expectation for a fixed feature matrix $\boldsymbol{F}$ and $\mathds{1}_{\mathcal{A}}(\boldsymbol{F})$ is a indicator function (1 if $\F \in \mathcal{A}$). We refer to $\mathcal{A}$ as the admissible set of feature matrices defined in Appendix \ref{appendix:admissible_feature_matrices}. We also have $\mathbb{P}\left(\mathcal{A}^c\right)=o(1)$, proof of which is also given in Appendix \ref{appendix:admissible_feature_matrices}. Now, we need to show
\begin{align}
\begin{aligned}
& \sup _{\boldsymbol{F} \in \mathcal{A}}\left|\mathbb{E}_{\backslash \boldsymbol{F}} \varphi\left(\frac{1}{k} \Phi_{\boldsymbol{A}}\right)-\mathbb{E}_{\backslash \boldsymbol{F}} \varphi\left(\frac{1}{k} \Phi_{\boldsymbol{B}}\right)\right| \leq  \max \left\{\left\|\varphi^{\prime}\right\|_{\infty}, \left\|\varphi^{\prime \prime}\right\|_{\infty}\right\} \kappa_k,
\label{eq:lindeberg}
\end{aligned}
\end{align}
for some $\kappa_k = o(1)$. To find a useful bound for \eqref{eq:lindeberg}, we interpolate the path from $\Phi_\A$ to $\Phi_\B$ following Lindeberg's method \cite{lindeberg1922neue,korada2011applications}
\begin{align}
    \Phi_{t} &:= \min_{\w \in \R^k} \left\{ \sum_{i=1}^t \left(\w^T \b_i - y_i \right)^2 + \sum_{i=t+1}^m \left(\w^T \a_i - y_{i}\right)^2 + Q(\w) \right\},\\
    \hat{\w}_t &:= \argmin_{\w \in \R^k} \left\{ \sum_{i=1}^t \left(\w^T \b_i - y_i\right)^2 + \sum_{i=t+1}^m \left(\w^T \a_i - y_{i}\right)^2 +  Q(\w)  \right\}, \label{eq:w_t}
\end{align}
for $0 \leq t \leq m$. Then,
\begin{align} 
    \left| \mathbb{E}_{\backslash \boldsymbol{F}} \varphi \left( \frac{1}{k} \Phi_\A \right) - \mathbb{E}_{\backslash \boldsymbol{F}} \varphi \left( \frac{1}{k} \Phi_\B \right) \right| \leq \sum_{t=1}^{m} 
    \left| \mathbb{E}_{\backslash \boldsymbol{F}} \varphi \left( \frac{1}{k} \Phi_{t} \right) - \mathbb{E}_{\backslash \boldsymbol{F}} \varphi \left( \frac{1}{k} \Phi_{t-1} \right) \right|, \label{eq:lindeberg_sum}
\end{align}
due to triangle inequality. Therefore, we focus on showing the following
\begin{align}
    \left| \mathbb{E}_{\backslash \boldsymbol{F}} \varphi \left( \frac{1}{k} \Phi_{t} \right) - \mathbb{E}_{\backslash \boldsymbol{F}} \varphi \left( \frac{1}{k} \Phi_{t-1} \right) \right| \leq \max \left\{\left\|\varphi^{\prime}\right\|_{\infty}, \left\|\varphi^{\prime \prime}\right\|_{\infty}\right\} \frac{\kappa_k}{k}, \label{eq:lindeberg_step}
\end{align}
since $k/m \in (0,\infty)$. Here, $\Phi_t$ and $\Phi_{t-1}$ can be seen as perturbations of a common "leave-one-out" problem defined as
\begin{align}
\Phi_{\backslash t} &:= \min_{\w \in \R^k} \left\{ \sum_{i=1}^{t-1} \left(\w^T \b_i - y_{i}\right)^2 + \sum_{i=t+1}^m \left(\w^T \a_i - y_{i}\right)^2 + Q(\w) \right\}, \label{eq:phi_t}\\
\hat{\w}_{\backslash t} &:= \argmin_{\w \in \R^k} \left\{ \sum_{i=1}^{t-1} \left(\w^T \b_i - y_{i}\right)^2 + \sum_{i=t+1}^m \left(\w^T \a_i - y_{i}\right)^2 + Q(\w) \right\}. \label{eq:hat_w}
\end{align}
Note that $t$-th sample is left out in the definition of $\Phi_{\backslash t}$ and $\w_{\backslash t}$. Since $\Phi_t \approx \Phi_{\backslash t}$, we apply Taylor's expansion around $\Phi_{\backslash t}$
\begin{align}
\begin{gathered}
\varphi\left(\frac{1}{k} \Phi_t\right)=\varphi\left(\frac{1}{k} \Phi_{\backslash t}\right)+\frac{1}{k} \varphi^{\prime}\left(\frac{1}{k} \Phi_{\backslash t}\right)\left(\Phi_t-\Phi_{\backslash t}\right) +\frac{1}{2 k^2} \varphi^{\prime \prime}(\zeta)\left(\Phi_t-\Phi_{\backslash t}\right)^2,
\end{gathered}
\end{align}
where $\zeta$ is some value that lies between $\frac{1}{k} \Phi_t$ and $\frac{1}{k} \Phi_{\backslash t}$. Similarly, applying Taylor's expansion to $\varphi\left(\Phi_{t-1}\right)$ around $\Phi_{\backslash t}$, and then subtracting it from the last equation, we get
\begin{align}
&\Bigg|\mathbb{E}_{\backslash \boldsymbol{F}}  \left[\varphi\left(\frac{1}{k} \Phi_t\right)\right]-\mathbb{E}_{\backslash \boldsymbol{F}}\left[\varphi\left(\frac{1}{k} \Phi_{t-1}\right)\right]\Bigg| \nonumber\\
&\leq  \frac{\left\|\varphi^{\prime}(x)\right\|_{\infty}}{k} \mathbb{E}_{\backslash \boldsymbol{F}} \left|\mathbb{E}_t\left(\Phi_t-\Phi_{t-1}\right)\right| + \frac{\left\|\varphi^{\prime \prime}(x)\right\|_{\infty}}{2 k}\frac{\left(\mathbb{E}_{\backslash \boldsymbol{F}}\left(\Phi_t-\Phi_{\backslash t}\right)^2+\mathbb{E}_{\backslash \boldsymbol{F}}\left(\Phi_{t-1}-\Phi_{\backslash t}\right)^2\right)}{k}, \label{eq:leave-one-out-taylor-expansion}
\end{align}
where $\mathbb{E}_t$ denotes the conditional expectation over the random vectors $\left\{\boldsymbol{a}_t, \boldsymbol{b}_t\right\}$ associated with the $t$-th training sample, while $\left\{\boldsymbol{a}_i, \boldsymbol{b}_i\right\}_{i \neq t}$ and $\boldsymbol{F}$ are fixed.

To bound the terms on the right side of \eqref{eq:leave-one-out-taylor-expansion}, we first define a surrogate optimization problem:
\begin{align}
& \Psi_t(\br) := \Phi_{\backslash t}+\min _{\w \in \mathbb{R}^k} \left\{ \frac{1}{2}\left(\w-\hat{\w}_{\backslash t}\right)^{T} \boldsymbol{H}_{\backslash t}\left(\w-\hat{\w}_{\backslash t}\right) + \left(\w^T \br - y_t \right)^2 \right\},
\label{eq:psi_t}
\end{align}
where 
\begin{align}
\boldsymbol{H}_{\backslash t} := & 2 \sum_{i=1}^{t-1} \b_i \b_i^T + 2 \sum_{i=t+1}^m \a_i \a_i^T + \nabla^2 Q\left(\hat{\w}_{\backslash t}\right)
\label{eq:H_t}
\end{align}
is the the Hessian matrix of the objective function $\Phi_{\backslash t}$ evaluated at $\hat{\w}_{\backslash t}$. 
Next, we have the following Lemma proven in Supplementary Lemma \ref{supplement_lemma:equality_of_surrogate}, which makes $\Psi_t(\br)$ particularly interesting.
\begin{lemma}
    $\Phi_{t-1} = \Psi_t(\a_t)$, and $\Phi_{t} = \Psi_t(\b_t)$ hold in this setting.
    \label{lemma:equality_of_surrogate}
\end{lemma}

We continue by simplifying $\Psi_t(\br)$ as follows:
\begin{align}
    \Psi_t(\br) :&= \Phi_{\backslash t}+\min _{\w \in \mathbb{R}^k} \left\{ \frac{1}{2}\left(\w-\hat{\w}_{\backslash t}\right)^{T} \boldsymbol{H}_{\backslash t}\left(\w-\hat{\w}_{\backslash t}\right) + \left(\w^T \br - y_t \right)^2 \right\},\\
    &= \Phi_{\backslash t}+\min_{\tau} \min_{\br^T (\w - \hat{\w}_{\backslash t}) = \tau} \left\{ \frac{1}{2}\left(\w-\hat{\w}_{\backslash t}\right)^{T} \boldsymbol{H}_{\backslash t}\left(\w-\hat{\w}_{\backslash t}\right) + \left(\hat{\w}_{\backslash t}^T \br + \tau - y_t \right)^2 \right\},\\
    &= \Phi_{\backslash t}+\min_{\tau} \left\{ \frac{\tau^2}{2\nu_t(\br)} + \left(\hat{\w}_{\backslash t}^T \br + \tau - y_t \right)^2 \right\}, \quad \text{ with } \quad \nu_t(\br) := \br^{T} \boldsymbol{H}_{\backslash t}^{-1} \br, \label{eq:moreau_envelope}\\
    &= \Phi_{\backslash t}+ \frac{\left(\hat{\w}_{\backslash t}^T \br - y_t \right)^2}{2\nu_t(\br) + 1} \leq \Phi_{\backslash t}+ \left(\hat{\w}_{\backslash t}^T \br - y_t \right)^2,
    \label{eq:surrogate_vs_loss}
\end{align}
where the inequality in the last line is achieved by using $\tau = 0$ in \eqref{eq:moreau_envelope} while the rest of the steps are trivial. Next, as shown in Supplementary Lemma \ref{supplement_lemma:surrogate_vs_leave_one_out}, we have
\begin{align}
\max \left\{\mathbb{E}_{\backslash \boldsymbol{F}}\left(\Psi_t\left(\boldsymbol{b}_t\right)-\Phi_{\backslash t}\right)^2, \mathbb{E}_{\backslash \boldsymbol{F}}\left(\Psi_t\left(\boldsymbol{a}_t\right)-\Phi_{\backslash t}\right)^2\right\} 
&\leq k^{1-\epsilon} \text{ polylog } k,
\end{align}
which holds uniformly over $\F \in \mathcal{A}$ and $t \in \{1,2,\dots,m\}$ for some $\epsilon > 0$ satisfying $\theta^2 \leq n^{1-\epsilon}$.
Then, we can reach
\begin{align}
\frac{1}{k}\mathbb{E}_{\backslash \boldsymbol{F}}\left(\Phi_t-\Phi_{\backslash t}\right)^2 
&= \frac{1}{k}\mathbb{E}_{\backslash \boldsymbol{F}}\left(\Psi_t\left(\boldsymbol{b}_t\right)-\Phi_{\backslash t}\right)^2 = o(1),
\label{eq:lindeberg_step_bound}
\end{align}
and similarly,
\begin{align}
\frac{1}{k}\mathbb{E}_{\backslash \boldsymbol{F}}\left(\Phi_{t-1}-\Phi_{\backslash t}\right)^2 &= \frac{1}{k}\mathbb{E}_{\backslash \boldsymbol{F}}\left(\Psi_t\left(\boldsymbol{a}_t\right)-\Phi_{\backslash t}\right)^2 = o(1).
\end{align}

Next, by Lemma \ref{lemma:equality_of_surrogate} and equation \eqref{eq:moreau_envelope}, we have
\begin{align}
    \mathbb{E}&_{\backslash \boldsymbol{F}}\left|\mathbb{E}_k\left(\Phi_t-\Phi_{t-1}\right)\right| = \mathbb{E}_{\backslash \boldsymbol{F}}\left|\mathbb{E}_t\left[\Psi_t\left(\boldsymbol{b}_t\right)-\Psi_t\left(\boldsymbol{a}_t\right)\right]\right|, \\
    &= \mathbb{E}_{\backslash \boldsymbol{F}}\left|\mathbb{E}_t\left[\frac{\left(\hat{\w}_{\backslash t}^T \b_t - y_t \right)^2}{2\nu_t(\b_t) + 1}-\frac{\left(\hat{\w}_{\backslash t}^T \a_t - y_t \right)^2}{2\nu_t(\a_t) + 1}\right]\right|,\\
    &\leq \frac{1}{2 \bar{\nu}_t +1} \mathbb{E}_{\backslash \boldsymbol{F}}\left|\mathbb{E}_t\left[\left(\hat{\w}_{\backslash t}^T \b_t - y_t \right)^2-\left(\hat{\w}_{\backslash t}^T \a_t - y_t \right)^2\right]\right| + \Delta_t(\a_t) + \Delta_t(\b_t), \label{eq:delta_decomposition}
\end{align}
where $\bar{\nu}_t := \E_t \nu_t(\a_t)$ and $\Delta_t(\br)$ is defined as 
\begin{align}
    \Delta_t(\br) := \mathbb{E}_{\backslash \boldsymbol{F}}\left| \mathbb{E}_t \left(\frac{1}{2\nu_t(\br) + 1} - \frac{1}{2\bar{\nu}_t + 1} \right) \left(\hat{\w}_{\backslash t}^T \br - y_t \right)^2 \right|,
\end{align}
for which the following is expected to hold
\begin{align}
    \max \left\{\Delta_t(\a_t), \Delta_t(\b_t)\right\} = o(1),\label{eq:delta_bound}
\end{align}
due to the concentration of $\nu_t(\a_t)$ and $\nu_t(\b_t)$ around $\bar{\nu}_t$  (Lemma \ref{lemma:nu_tail_bound}) and the fact that $\frac{1}{2 \bar{\nu}_t +1} \leq C$ for some $C > 0$ by \eqref{eq:surrogate_vs_loss}.

\begin{lemma}
    Suppose the definitions in Appendix \ref{appendix:equivalence_proof}. Let $\epsilon >0$ be a constant satisfying $\theta \leq n^{1/2-\epsilon/2}$. For $\br_t = \a_t$ or $\br_t = \b_t$, we have
    \begin{align}
        \Prob(|  \nu_t(\br_t) - \E_t [\nu_t(\br_t)] | > \delta) \leq c e^{-Ck^{\epsilon} \delta^2 / \text{polylog } k}, \label{eq:nu_t_tail_bound}
    \end{align}
     for some $c,C >0$. Furthermore,
    \begin{align}
        \E_t[| \nu_t(\br_t) - \E_t [\nu_t(\br_t)]|^l] \leq 2 c (l!) (C k^{\epsilon} / \text{polylog } k)^{-l/2}, \label{eq:nu_t_moment_bound}
    \end{align}
    for some $c,C >0$ and any $l \in \mathbb{Z}^+$. Finally,
    \begin{align}
        \left|\E_t \nu_t(\a_t) - \E_t [\nu_t(\b_t)]\right| = \mathcal{O}(k^{-\varsigma}), \label{eq:nu_a_t_vs_nu_b_t}
    \end{align}
    for some $\varsigma >0$.
    \begin{proof}
        Supplementary Lemma \ref{supplement_lemma:nu_tail_bound}
    \end{proof}
    \label{lemma:nu_tail_bound}
\end{lemma}

In bounding the remaining term, the following lemma (proven in Supplementary Lemma \ref{supplement_lemma:w_t_bound}) will be useful. 

\begin{lemma}
    For any $l \in \mathbb{Z}^+$ and any $t \in \{0,\dots,m\}$, we have $\mathbb{E}_{\backslash \F}  \|\hat{\w}_t\|^l \leq \text{polylog } k$, and $\mathbb{E}_{\backslash \F}  \|\hat{\w}_{\backslash t}\|^l \leq \text{polylog } k$.
    \label{lemma:w_t_bound}
\end{lemma}

Next, we bound the remaining term in \eqref{eq:delta_decomposition} as 
\begin{align}
    \mathbb{E}_{\backslash \boldsymbol{F}}&\left|\mathbb{E}_t\left[\left(\hat{\w}_{\backslash t}^T \b_t - y_t \right)^2 \right] - \mathbb{E}_t\left[\left(\hat{\w}_{\backslash t}^T \a_t - y_t \right)^2\right]\right|,\\
    &= \mathbb{E}_{\backslash \boldsymbol{F}}\left|\hat{\w}_{\backslash t}^T \left(\E_t[\b_t \b_t^T] - \E_t[\a_t \a_t^T]\right) \hat{\w}_{\backslash t} - 2 \hat{\w}_{\backslash t}^T \left(\E_t[\b_t y_t] - \E_t[\a_t y_t]\right) \right|\\
    &\stackrel{(a)}{\leq} \mathbb{E}_{\backslash \boldsymbol{F}}\left|\hat{\w}_{\backslash t}^T \left(\E_t[\b_t \b_t^T] - \E_t[\a_t \a_t^T]\right) \hat{\w}_{\backslash t}\right| + 2 \mathbb{E}_{\backslash \boldsymbol{F}}\left| \hat{\w}_{\backslash t}^T \left(\E_t[\b_t y_t] - \E_t[\a_t y_t]\right) \right|\\
    &\stackrel{(b)}{\leq} \mathbb{E}_{\backslash \boldsymbol{F}} \left\|\hat{\w}_{\backslash t}\right\|^2 \mathbb{E}_{\backslash \boldsymbol{F}} \left\|\E_t[\b_t \b_t^T] - \E_t[\a_t \a_t^T]\right\|  + 2\mathbb{E}_{\backslash \boldsymbol{F}} \left\|\hat{\w}_{\backslash t}\right\| \mathbb{E}_{\backslash \boldsymbol{F}}\left\|\E_t[\b_t y_t] - \E_t[\a_t y_t]\right\| \\
    &\stackrel{(c)}{=} o(1),
    \label{eq:surrogate_difference}
\end{align}
where we use triangle inequality to reach (a). Then, we apply Cauchy-Schwarz inequality in order to reach (b). In the last step, we use a bound on $\mathbb{E}_{\backslash \boldsymbol{F}}\left\|\hat{\w}_{\backslash t}\right\|^l$ (Lemma \ref{lemma:w_t_bound}) with the equivalence of covariance matrices \eqref{eq:Sigma_x} and the equivalence of cross-covariance vectors \eqref{eq:Sigma_xy} to reach (c). We can use \eqref{eq:surrogate_difference} and \eqref{eq:delta_bound} to bound the right-hand side of \eqref{eq:delta_decomposition}. Then, \eqref{eq:lindeberg_step_bound}-\eqref{eq:delta_decomposition} together with \eqref{eq:leave-one-out-taylor-expansion} imply \eqref{eq:lindeberg_step}. Finally, \eqref{eq:lindeberg_sum}-\eqref{eq:lindeberg_step} imply \eqref{eq:lindeberg} and consequently \eqref{eq:test_difference}, which completes our proof. 

\section{Admissible Set of Feature Matrices}
\label{appendix:admissible_feature_matrices}
A feature matrix $\F = [\f_1, \f_2, \dots, \f_k]^T$ is called admissible ($\F \in \mathcal{A}$) if it satisfies
\begin{align}
    \max _{1 \leq i, j \leq k}\left|\f_i^T \f_j +  \theta \bgamma^T \f_j \f_i^T \bgamma  - \delta_{i j}\right| &\leq \frac{\text{polylog } k}{\sqrt{k}},\label{eq:f_cond1}\\
    \|\F (\I_n + (\sqrt{1+\theta} - 1) \bgamma\bgamma^T)\| & \leq \sqrt{\theta} \text { 
 polylog } k, \label{eq:f_cond2}
\end{align}
where $\delta_{i j}$ denote the Kronecker delta. Note that \eqref{eq:f_cond1} (see Lemma \ref{lemma:f_cond1}) and \eqref{eq:f_cond2} (see Lemma \ref{lemma:f_cond2}) hold with high probability for $\f_i \sim \Normal\left(0, \frac{1}{n+\theta} \I_n\right)$ which is sampled independent of $\bgamma$, $\bxi$ and $\f_j$ for $i \neq j$. 

\begin{lemma}
    Under assumptions S.1-S.6, the following holds with high probability 
    \begin{align}
        \max _{1 \leq i < j \leq k}\left|\f_i^T \f_j +  \theta \bgamma^T \f_j \f_i^T \bgamma \right| &\leq \frac{\text{polylog } k}{\sqrt{k}},\\
        \max _{1 \leq i \leq k}\left|\|\f_i\|^2 +  \theta (\bgamma^T \f_i)^2  - 1\right| &\leq \frac{\text{polylog } k}{\sqrt{k}}.
    \end{align}
    \begin{proof}
        We first apply triangle equality to reach
        \begin{align}
            \left|\f_i^T \f_j +  \theta \bgamma^T \f_j \f_i^T \bgamma \right| &\leq |\f_i^T \f_j| + \theta |\bgamma^T \f_j \f_i^T \bgamma |, \label{eq:f_cond_1_1_terms}\\
            \left|\|\f_i\|^2 +  \theta (\bgamma^T \f_i)^2 - 1 \right| &\leq \left|\|\f_i\|^2 - \frac{n}{n+\theta} \right| +  \theta \left|(\bgamma^T \f_i)^2 - \frac{1}{n+\theta} \right|. \label{eq:f_cond_1_2_terms}
        \end{align}
        Next, we find probability bounds for each of the terms on the right-hand side of \eqref{eq:f_cond_1_1_terms} and \eqref{eq:f_cond_1_2_terms}. Let's start with $|\f_i^T \f_j|$. Remember that $\f_i \sim \Normal\left(0, \frac{1}{n+\theta} \I_n\right)$. Then, for any $l \in \{1,\dots,n\}$, $f_{i,l}$ and $f_{j,l}$ are sub-Gaussian random variables with sub-Gaussian norm bounded by $\frac{C}{\sqrt{n+\theta}}$ for some $C > 0$ \cite[Example 2.5.8]{vershynin2018high}. Therefore, $f_{i,l} f_{j,l}$ is a sub-exponential random variable with sub-exponential norm bounded by $\frac{C^2}{n+\theta}$ \cite[Lemma 2.7.7]{vershynin2018high}. Since $\f_i^T \f_j = \sum_{l=1}^{n} f_{i,l} f_{j,l}$, we can apply Bernstein's inequality \cite[Theorem 2.8.2]{vershynin2018high} 
        \begin{align}
            \Prob \left( |\f_i^T \f_j| \geq \epsilon \right) \leq 2 e^{-c \min \left(\frac{\epsilon^2}{K^2 n}, \frac{\epsilon}{K} \right)},
        \end{align}
        where $K = C^2/(n+\theta)$ is the sub-exponential norm bound and $c > 0$ is a constant. We set $\epsilon = (\log k)^2/\sqrt{k}$ and apply union bound so that we get
        \begin{align}
            \Prob \left( \max _{1 \leq i < j \leq k} |\f_i^T \f_j| \geq \frac{(log k)^2}{\sqrt{k}} \right) \leq c_1 e^{-(\log k)^2 / c_2},
        \end{align}
        for some $c_1, c_2 > 0$. Note that we use $n/k \in (0,\infty)$ and $\theta \leq \sqrt{n}$ to write $n$ and $\theta$ in terms of $k$. Also, we may reach the following inequality using the analogous steps:
        \begin{align}
             \Prob \left( \max _{1 \leq i \leq k} \left|\|\f_i\|^2 - \frac{n}{n+\theta} \right| \geq \frac{(log k)^2}{\sqrt{k}} \right) \leq c_1 e^{-(\log k)^2 / c_2 },
        \end{align}
        for some $c_1, c_2 > 0$. Note that $n/(n+\theta)$ appears in the equation since $\E [\|\f_i\|^2] = n/(n+\theta)$.
        Now, we focus on $|\bgamma^T \f_j \f_i^T \bgamma|$. First, note that $(\f_i^T \bgamma) \sim \Normal(0, 1/(n+\theta))$ since $\bgamma$ is a deterministic vector with $\|\bgamma\| = 1$. Therefore, $(\f_i^T \bgamma)$ and $(\f_j^T \bgamma)$ are sub-Gaussian random variables with sub-Gaussian norm bounded by $\frac{C}{\sqrt{n+\theta}}$ for some $C > 0$ \cite[Example 2.5.8]{vershynin2018high} so $(\bgamma^T \f_j \f_i^T \bgamma)$ is a sub-exponential random variable with sub-exponential norm bounded by $\frac{C^2}{n+\theta}$ \cite[Lemma 2.7.7]{vershynin2018high}. Thus, we have sub-exponential tail bound \cite[Proposition 2.7.1]{vershynin2018high} as follows
        \begin{align}
            \Prob \left( |\bgamma^T \f_j \f_i^T \bgamma| \geq \epsilon \right) \leq 2 e^{-C\epsilon (n+\theta)},
        \end{align}
        for some $C > 0$. Setting $\epsilon = (\log k)^2/k$, we reach
        \begin{align}
            \Prob \left( \max _{1 \leq i < j \leq k} |\bgamma^T \f_j \f_i^T \bgamma| \geq \frac{(log k)^2}{k} \right) \leq 2 e^{-C (\log k)^2},
            \label{eq:f_gamma_bound}
        \end{align}
        for some $C > 0$. Following the analogous steps, we also get
        \begin{align}
            \Prob \left( \max _{1 \leq i \leq k} \left|(\f_i^T \bgamma)^2 - \frac{1}{n+\theta} \right| \geq \frac{(log k)^2}{k} \right) \leq 2 e^{-C(\log k)^2},
            \label{eq:f_gamma_2_bound}
        \end{align}
        for some $C > 0$ since $\E[(\f_i^T \bgamma)^2] = 1/(n+\theta)$. Using the found probability bounds for each of the terms on the right-hand side of \eqref{eq:f_cond_1_1_terms} and \eqref{eq:f_cond_1_2_terms}, we complete the proof.
    \end{proof}
    \label{lemma:f_cond1}
\end{lemma}

\begin{lemma}
    Under assumptions S.1-S.6, the following hold with high probability
    \begin{equation}
        \|\F (\I_n + \theta \bgamma \bgamma^T) \F^T \| \leq \theta \text{ polylog } k.
    \end{equation}
    \begin{proof}
        Using the triangle inequality, we get
        \begin{equation}
             \|\F (\I_n + \theta \bgamma \bgamma^T) \F^T \| \leq \|\F \F^T \| +  \theta \|\F \bgamma \bgamma^T \F^T \|.
             \label{eq:f_cond2_triangle}
        \end{equation}
        First, we focus on $\|\F \F^T\| = \|\F\|^2$. Here, we can take advantage of a well-known result on the spectral norm of Gaussian random matrices \cite[Corollary 7.3.3]{vershynin2018high}: 
        \begin{align}
            \Prob \left( \|\F\| \geq \sqrt{\frac{n}{n+\theta}} + \sqrt{\frac{k}{n+\theta}} + \epsilon \right) &\leq 2 e^{- c (n+\theta) \epsilon^2},\\
            \Prob \left( \|\F\| \geq \sqrt{\frac{n}{n+\theta}} + 2\sqrt{\frac{k}{n+\theta}}  \right) &\leq 2 e^{- c k}
        \end{align}
        for any $\epsilon > 0$ and some $c > 0$. This result indicates $\|\F \F^T\| \leq \text{polylog } k$ with high probability.
        Next, we work on $\|\F \bgamma \bgamma^T \F^T \|$. Since $(\F \bgamma \bgamma^T \F^T )_{i,j} = \f_i^T \bgamma \bgamma^T \f_j$, we may use the related results \eqref{eq:f_gamma_bound} - \eqref{eq:f_gamma_2_bound} from the proof of Lemma \ref{lemma:f_cond1}. Specifically, we have the following inequality with high probability
        \begin{align}
            \max _{1 \leq i, j \leq k}\left| \bgamma^T \f_j \f_i^T \bgamma  - \frac{\delta_{i j}}{n+\theta} \right| &\leq \frac{\text{polylog } k}{k},\\
            \max _{1 \leq i, j \leq k}\left| \bgamma^T \f_j \f_i^T \bgamma  \right|  &\leq \frac{\text{polylog } k}{k},
        \end{align}
        where $\delta_{i j}$ denote the Kronecker delta function and. we use $1/(n+\theta) < C/k$ for some $C > 0$ in reaching the last line. It follows that the following holds with high probability
        \begin{equation}
            \|\F \bgamma \bgamma^T \F^T \| \leq \|\F \bgamma \bgamma^T \F^T \|_F  = \sqrt{\sum_{i=1}^{k}\sum_{j=1}^{k} (\f_i^T \bgamma \bgamma^T \f_j)^2 } \leq \text{polylog } k.
        \end{equation}
        Using \eqref{eq:f_cond2_triangle}, we combine the results and reach the statement of the lemma.
    \end{proof}
    \label{lemma:f_cond2}
\end{lemma}

\section{Equivalence of Covariance Matrices} \label{appendix:covariance_equivalence}
Here, we show that \eqref{eq:Sigma_x} holds for any activation function $\sigma$ satisfying assumption S.6 and a second activation function defined as $\hat{\sigma}(x) := \mu_1 x + \mu_* z$ where $z \sim \Normal(0,1)$. Note that trivially we have $\E_\x [\hat{\sigma}(\F \x) \hat{\sigma}(\F \x)^T] = \mu_1^2 \F (\I_n + \theta \bgamma \bgamma^T \F^T) + \mu_*^2 \I_k$. The following lemma shows the equivalence of covariance matrices for the activation functions $\sigma$ and $\hat{\sigma}$.

\begin{lemma}
    Under assumptions S.1-S.6 given in Section \ref{sec:assumptions}, the following holds
    \begin{align}
        \left\|\mathbb{E}_{\x}[\sigma(\F\x) \sigma(\F \x)^T] - (\mu_1^2 \hat{\F} \hat{\F}^T + \mu_*^2 \I_k)\right\| = \mathcal{O}(k^{-\varsigma}),
    \end{align}
    for some $\varsigma >0$,
    where $\hat{\F} := \F (\I_n + (\sqrt{1+\theta} - 1) \bgamma\bgamma^T)$.
    \begin{proof}
        \cite[Lemma 5]{hu2022universality} showed 
$\|\E_\g[\sigma(\F\g) \sigma(\F \g)^T] - (\mu_1^2 \F \F^T + \mu_*^2 \I_k)\| = \mathcal{O}(k^{-\varsigma})$ for some $\varsigma >0$ under $\g \sim \Normal(0, \I_n)$ and some other assumptions on $\F$ matrix and activation function $\sigma$. Here, we adapt their proof to our case. First, note that we can consider $\sigma(\hat{\F}\g)$ instead of $\sigma(\F \x)$ since $\sigma(\hat{\F}\g)$ is equal (in distribution) to $\sigma(\F \x)$. Now, our goal is to show $\|\E_\g[\sigma(\hat{\F}\g) \sigma(\hat{\F} \g)^T] - (\mu_1^2 \hat{\F} \hat{\F}^T + \mu_*^2 \I_k)\| = \mathcal{O}(k^{-\varsigma})$ for some $\varsigma >0$. The $(i, j)$-th entry of $\E_\g[\sigma(\hat{\F}\g) \sigma(\hat{\F} \x)^T]$ is $\mathbb{E}\left[\sigma\left(\hat{\f}_i^T \g\right) \sigma\left(\hat{\f}_j^T \g\right)\right]$. Since $(\hat{\f}_i^T \g, \hat{\f}_j^T \g)$ are jointly Gaussian, we can rewrite their joint distribution as that of $\left(z_i, \rho_{i j} z_i+\sqrt{1-\rho_{i j} \rho_{j i}} z_j\right)$, where $z_i \sim \mathcal{N}(0,\|\hat{\f}_i\|^2)$, $z_j \sim \mathcal{N}(0,\|\hat{\f}_j\|^2)$ are two independent Gaussian variables and $\rho_{i j} := \hat{\f}_i^T \hat{\f}_j / \|\hat{\f}_i\|^2$. Then, for $i \neq j$, we have,
\begin{align}
\mathbb{E}&\left[\sigma\left(\hat{\f}_i^T \g\right) \sigma\left(\hat{\f}_j^T \g\right)\right] = \mathbb{E}\left[\sigma\left(z_i\right) \sigma\left(\rho_{i j} z_i+\sqrt{1-\rho_{i j} \rho_{j i}} z_j\right)\right], \\
&\stackrel{(a)}{=} \mathbb{E}[\sigma\left(z_i\right)  \sigma\left(\sqrt{1-\rho_{i j} \rho_{j i}} z_j\right)] +\rho_{i j} \mathbb{E}\left[\sigma\left(z_i\right) z_i \sigma^{\prime}\left(\sqrt{1-\rho_{i j} \rho_{j i}} z_j\right)\right] \nonumber\\
&\qquad+\frac{1}{2} \rho_{i j}^2 \mathbb{E}\left[\sigma\left(z_i\right) z_i^2 \sigma^{\prime \prime}\left(\sqrt{1-\rho_{i j} \rho_{j i}} z_j\right)\right] 
 +\frac{1}{6} \rho_{i j}^3 \mathbb{E}\left[\sigma\left(z_i\right) z_i^3 \sigma^{\prime \prime \prime}\left(\zeta_{i j}\right)\right], \\
&\stackrel{(b)}{=}\hat{\f}_i^T \hat{\f}_j\mathbb{E} \sigma^{\prime}\left(z_i\right) \mathbb{E} \sigma^{\prime}\left(\sqrt{1-\rho_{i j} \rho_{j i}} z_j\right) + \frac{1}{6} \rho_{i j}^3 \mathbb{E}\left[\sigma\left(z_i\right) z_i^3 \sigma^{\prime \prime \prime}\left(\zeta_{i j}\right)\right], \\
&\stackrel{(c)}{=}\hat{\f}_i^T \hat{\f}_j \mathbb{E} \sigma^{\prime}\left(z_i\right) \mathbb{E} \sigma^{\prime}\left(z_j\right)+R_{i j}, \label{eq:covariance_structure}
\end{align}
where $\zeta_{i j}$ is some point between $\sqrt{1-\rho_{i j} \rho_{j i}} z_j$ and $\rho_{i j} z_i+ \sqrt{1-\rho_{i j} \rho_{j i}} z_j$. Here, we apply Taylor's series expansion of $\sigma(\rho_{i j} z_i+\sqrt{1-\rho_{i j} \rho_{j i}} z_j)$ around $\sqrt{1-\rho_{i j} \rho_{j i}} z_j$ to reach (a). Then, we use the independence between $z_i$ and $z_j$, and the following identities: $\mathbb{E} \sigma(z_i) = \mathbb{E} [\sigma(z_i) z_i^2] =0$ (due to $\sigma$ being an odd function) and $\mathbb{E}[\sigma(z_i)  z_i]=\|\hat{\f}_i\|^2 \mathbb{E}[\sigma^{\prime}(z_i)]$ (by Stein's lemma). To reach (c), we define the remainder term $R_{i j}$ as
\begin{align}
R_{i j}= \hat{\f}_i^T \hat{\f}_j\mathbb{E} \sigma^{\prime}\left(z_i\right)\left(\mathbb{E} \sigma^{\prime}\left(\sqrt{1-\rho_{i j} \rho_{j i}} z_j\right)-\mathbb{E} \sigma^{\prime}\left(z_j\right)\right) + \frac{1}{6} \rho_{i j}^3 \mathbb{E}\left[\sigma\left(z_i\right) z_i^3 \sigma^{\prime \prime \prime}\left(\zeta_{i j}\right)\right].
\label{eq:R_ij}
\end{align}

For $i=j$, we define $R_{i i}=0$. Then, we can verify the following decomposition using \eqref{eq:covariance_structure}:
\begin{align}
\E[\sigma(\hat{\F}\g) \sigma(\hat{\F} \g)^T] = \left(\mu_1 \I_k +\D_1\right) \hat{\F} \hat{\F}^T\left(\mu_1 \I_k+\D_1\right)+\mu_{*}^2 \I_k+\D_2+\D_3+\mathbf{R}
\end{align}
where we define $\D_1, \D_2, \D_3$ to be diagonal matrices as follows: $\D_1:=\text{diag }(\mathbb{E} \sigma^{\prime}\left(z_i\right)-\mu_1)$, $\D_2=\text{diag }(\mu_1^2 - \| \hat{\f}_i \|^2(\mathbb{E} \sigma^{\prime}(z_i))^2)$, and $\D_3=\text{diag }(\mathbb{E} \sigma^2(z_i)-\mu_1^2-\mu_{*}^2)$.

Then, we have
\begin{align}
\left\|\E[\sigma(\F\x) \sigma(\F \x)^T] - (\mu_1^2 \hat{\F} \hat{\F}^T + \mu_*^2 \I_k)\right\| &\leq  
\left(2 \mu_1+\right. \left.\left\|\mathbf{D}_1\right\|\right)\|\hat{\F}\|^2\left\|\mathbf{D}_1\right\|+\left\|\mathbf{D}_2\right\|+\left\|\mathbf{D}_3\right\| + \left\|\mathbf{R}\right\|,\\
&\leq \frac{\|\hat{\mathbf{F}}\|^2 \text{ polylog} k}{\sqrt{k}} = \mathcal{O}(k^{-\varsigma}),
\end{align}
for some $\varsigma >0$, where Lemma \ref{lemma:covariance_remainder_bounds} is used for the bounds of $\left\|\mathbf{D}_1\right\|,\left\|\mathbf{D}_2\right\|,\left\|\mathbf{D}_3\right\|$ and $\|\mathbf{R}\|$. To reach the conclusion, we then take advantage of $\|\hat{\F}\|^2 \leq k^{1/2 - \epsilon/2} \text{ polylog } k$ (for some $\epsilon >0$ satisfying $\theta \leq n^{1/2-\epsilon/2}$), which is due to \eqref{eq:f_cond2}. 
    \end{proof}
    \label{lemma:covariance_equivalence}
\end{lemma}

The following lemmas are auxiliary results used in the proof of the lemma above.

\begin{lemma}
    Suppose the setting in Lemma \ref{lemma:covariance_equivalence} and the definitions in its proof. Then,
    \begin{align}
        \max \left\{ \|\D_1\|, \|\D_2\|, \|\D_3\|, \|\mathbf{R}\| \right\} \leq \frac{\text{polylog } k}{\sqrt{k}}.
    \end{align}
    \begin{proof}
        Before bounding the terms, we define a truncated version of the activation function $\sigma$ as
        \begin{equation}
            \sigma_{\epsilon}(x) := \begin{cases}
            \sigma(x), & \text{if } |x| < \epsilon,\\
            0, & \text{otherwise,}
        \end{cases}
        \end{equation}
        for $\epsilon = 2log k$, which is useful due to Lemma \ref{lemma:truncated_expectation}. Furthermore, $\sigma_{\epsilon}(x)$, its finite powers and its derivatives are bounded due to assumption S.6.
        First, we focus on $\D_1$
        \begin{align}
            \|\D_1\| &\leq \max_i \mathbb{E} | \sigma^{\prime}\left(z_i\right)-\mu_1 |\\
            &\leq \max_i \mathbb{E}_{z \sim \Normal(0,1)} |\sigma^{\prime}(\|\hat{\f}_i\| z)- \sigma^{\prime}(z) |\\
            &\leq \max_i \mathbb{E}_{z \sim \Normal(0,1)} |\sigma_\epsilon^{\prime}(\|\hat{\f}_i\| z)- \sigma_\epsilon^{\prime}(z) | + \frac{\text{ polylog } k}{k^{log k}}  \label{eq:sigma_prime_truncation}\\
            &\leq \left( \|\sigma^{\prime\prime}_{\epsilon}(x)\|_\infty \E[|z|] \right) \max_i \left| \|\hat{\f}_i\| - 1 \right| + \frac{\text{ polylog } k}{k^{log k}}\\
            &\leq \left( \|\sigma^{\prime\prime}_{\epsilon}(x)\|_\infty \E[|z|] \right) \max_i \left| \|\hat{\f}_i\|^2 - 1 \right| + \frac{\text{ polylog } k}{k^{log k}} \leq \frac{\text{ polylog } k}{\sqrt{k}},
        \end{align}
        where we use Lemma \ref{lemma:truncated_expectation} to get \eqref{eq:sigma_prime_truncation} and while the last line is due to \eqref{eq:f_cond1} (Lemma \ref{lemma:f_cond1}) and bounded second derivative of $\sigma_\epsilon$. Similarly, one can show $\|\D_2\| \leq \text{ polylog } k / \sqrt{k}$. Next, we study $\|\D_3\|$ as
        \begin{align}
            \|\D_3\| &\leq \max_i  | \mathbb{E}\sigma^2\left(z_i\right)-\mu_1^2 -\mu_*^2 |\\
            &= \max_i \mathbb{E}_{z \sim \Normal(0,1)} | \sigma^2(\|\hat{\f}_i\| z) - \sigma^2(z) |\\
            &= \max_i \mathbb{E}_{z \sim \Normal(0,1)} | \sigma_{\epsilon}^2(\|\hat{\f}_i\| z) - \sigma_{\epsilon}^2(z) | + \frac{\text{ polylog } k}{k^{log k}}\\
            &\leq \text{ polylog } k \max_i \left| \|\hat{\f}_i\| - 1 \right| + \frac{\text{ polylog } k}{k^{log k}} \label{eq:sigma2_bounded_derivative}\\
            &\leq \text{ polylog } k \max_i \left| \|\hat{\f}_i\|^2 - 1 \right| + \frac{\text{ polylog } k}{k^{log k}} \leq \frac{\text{ polylog } k}{\sqrt{k}},
        \end{align}
        where we use the bounded derivative of $\sigma_{\epsilon}^2$ to reach \eqref{eq:sigma2_bounded_derivative} while the rest is similar to the bounding of $\|\D_1\|$. Lastly, to bound $\|\mathbf{R}\|$, one can first easily show that 
        \begin{equation}
            \max_{1 \leq i,j \leq k} | R_{i j} | \leq \frac{\text{polylog } k}{k\sqrt{k}}.
        \end{equation}
        using \eqref{eq:f_cond1} (Lemma \ref{lemma:f_cond1}) and the definition of $R_{i j}$ in \eqref{eq:R_ij}. Then, we have
        \begin{align}
            \| \mathbf{R} \| \leq \| \mathbf{R} \|_F = \left(\sum_{i=1}^k \sum_{j=1}^k R_{i j}^2 \right)^{1/2} \leq \frac{\text{ polylog } k}{\sqrt{k}},
        \end{align}
        which concludes our proof.
    \end{proof}
    \label{lemma:covariance_remainder_bounds}
\end{lemma}

\begin{lemma}
    Let $f: \R \to \R$ be a function satisfying $|f(x)| \leq C (1 + |x|^{K})$ for all $x \in \R$ and some constants $C > 0, K \in \mathbb{Z}^+$. Furthermore, define a truncated version of $f$ as
    \begin{align}
        f_{\epsilon}(x) := \begin{cases}
        f(x), & \text{if } |x| < \epsilon,\\
        0, & \text{otherwise,}
    \end{cases}
    \end{align}
    for $\epsilon > 0$. If $\epsilon =2log k$, the following holds for any constant $a > 0$ and $z \sim \Normal(0,1)$, 
    \begin{equation}
        \mathbb{E}_z|f(a z) - f_{\epsilon}(a z) | = \frac{\text{ polylog } k}{k^{log k}}.
    \end{equation}
    \begin{proof}
        Supplementary Lemma \ref{supplement_lemma:truncated_expectation}
    \end{proof}
    \label{lemma:truncated_expectation}
\end{lemma}

\section{Equivalence of Cross-Covariance Vectors}
\label{appendix:cross_covariance_equivalence}
Here, we are interested in the cross-covariance $\E[\sigma(\F\x) y]$. 

In showing the equivalence of the cross-covariance vectors, we use the following lemma describing the orthogonality of Hermite polynomials.

\begin{lemma}
    Let $H_i(x)$ be the probabilist's $i$-th Hermite polynomial and $z_1, z_2 \stackrel{i.i.d}{\sim} \Normal(0,1)$. Then, for any $\rho \in [0,1]$ and any $c \geq |\rho|$,
    \begin{align}
        \mathbb{E}_{z_1, z_2}[H_i(\rho z_1 + \sqrt{c^2- \rho^2} z_2) H_j(z_1)] = (i!) \rho^i \delta_{i,j},
    \end{align}
    where $\delta_{i j}$ is Kronecker delta function.
    \begin{proof}
        Supplementary Lemma \ref{supplement_lemma:hermite_correlated}
    \end{proof}
    \label{lemma:hermite_correlated}
\end{lemma}

The following lemma specifies the equivalence of the cross-covariance vectors for the activation function $\sigma$ and its finite-degree Hermite expansion $\hat{\sigma}_l$.

\begin{lemma}
    Suppose assumptions S.1-S.6. Let $\eta_i := \E \left[\frac{(\f_i\x) (\bxi^T \x)}{\sqrt{1+\theta\alpha^2}} \right] =\frac{\f_i^T\bxi + \theta \alpha \f_i^T\bgamma}{\sqrt{1+\theta\alpha^2}}$. Assume that  
    \begin{equation}
        \eta := \max _{1 \leq i \leq k} \left|\eta_i\right| \leq \frac{C}{n^{1/l}}, \quad \text{for some $C > 0$ and some $l \in \mathbb{Z}^+$.}
    \end{equation}
    Then, the following holds
    \begin{equation}
    \left\|\E_{(\x,y)} \left[\sigma(\F\x) y \right] - \E_{(\x,y)} \left[\hat{\sigma}_l(\F\x) y \right]\right\| = \mathcal{O}(k^{-\varsigma}), \label{eq:cross_covariance_bound}
\end{equation}
for some $\varsigma >0$ and $\hat{\sigma}_l(x)$ defined in \eqref{eq:equivalent_activation}.
\begin{proof}
    Let $\x \sim \Normal(0, \I_k+ \theta \bgamma \bgamma^T)$. By assumption S.3, we have
\begin{equation}
    \E_{(\x,y)} \left[\sigma(\F\x) y \right] = \E_{\x} \left[\sigma(\F\x) \sigma_*\left(\frac{\bxi^T \x}{\sqrt{1+\theta\alpha^2}}\right) \right] = \E_{\g} \left[\sigma \left(\hat{\F}\g \right) \sigma_*\left(\hat{\bxi}^T \g \right) \right],
\end{equation}
where $\g \sim \Normal(0, \I_n)$, $\hat{\F} := \F (\I_n + (\sqrt{1+\theta} - 1) \bgamma\bgamma^T)$ and $\hat{\bxi} := \frac{(\I_n + (\sqrt{1+\theta} - 1)\bgamma\bgamma^T) \bxi}{\sqrt{1+\theta \alpha^2}}$. Before dealing with the functions $\sigma$ and $\sigma_*$, we consider the joint distribution of $(\hat{\f}_i^T\g, \hat{\bxi}{}^T\g)$, which is jointly Gaussian
\begin{align}
    (\hat{\f}_i^T\g, \hat{\bxi}{}^T\g) \sim \Normal\left(0, \begin{bmatrix} \|\hat{\f}_i\|^2 &  \eta_{i} \\ \eta_{i}  & 1
\end{bmatrix}\right),
\end{align}
Indeed, we can write an equivalent of $(\hat{\f}_i^T\g, \hat{\bxi}{}^T\g)$ as $(\eta_i z + \sqrt{\|\hat{\f}_i\|^2 - \eta_i^2} z_i, z)$,
where $z_i,z \stackrel{i.i.d}{\sim} \Normal(0,1)$. This decomposition will be useful since we split correlated part $\eta_i z$ and uncorrelated part $\sqrt{\|\hat{\f}_i\|^2 - \eta_i^2} z_i$. Now, we use Hermite expansions of $\sigma$ and $\sigma_*$
\begin{align}
    \sigma(x) = \sum_{j=0}^{\infty} \frac{1}{j!} \mu_j H_j(x), \quad \text{ and } \quad \sigma_*(x) = \sum_{j=0}^{\infty} \frac{1}{j!} \Tilde{\mu}_j H_j(x),
\end{align}
where $H_j(z)$ is the probabilist's $j$-th Hermite polynomial, $\mu_j := \E_{z \sim \Normal(0,1)}[H_j(z) \sigma(z)]$ and $\Tilde{\mu}_j := \E_{z \sim \Normal(0,1)}[H_j(z) \sigma_*(z)]$.
Then, we have
\begin{align}
    \E_{(\x,y)} & \left[\sigma(\f_i^T\x) y \right] = \E_{\g}\left[\sigma(\hat{\f}_i^T\g)\sigma_*(\hat{\bxi}{}^T\g)\right],\\
    &= \E_{z_i, z}\left[\sigma\left(\eta_i z + \sqrt{\|\hat{\f}_i\|^2 - \eta_i^2} z_i\right)\sigma_*(z)\right]\\
    &= \E_{z_i,z}\left[\left( \sum_{j=0}^{\infty} \frac{1}{j!} \mu_j H_j\left(\eta_i z + \sqrt{\|\hat{\f}_i\|^2 - \eta_i^2} z_i\right) \right) \left( \sum_{j=0}^{\infty} \frac{1}{j!} \Tilde{\mu}_j H_j(z) \right) \right],\\
    &\stackrel{(a)}{=}  \sum_{j=0}^{\infty} \frac{1}{(j!)^2} \mu_j \Tilde{\mu}_j \E_{z_i,z}\left[H_j\left(\eta_i z + \sqrt{\|\hat{\f}_i\|^2 - \eta_i^2} z_i\right) H_j(z) \right] \\
    &\stackrel{(b)}{=}  \sum_{j=0}^{\infty} \frac{1}{j!} \mu_j \Tilde{\mu}_j \eta_i^j  \label{eq:Sigma_xy_decomposition} \stackrel{(c)}{=}  \left(\sum_{j=0}^{l-1} \frac{1}{j!} \mu_j \Tilde{\mu}_j \eta_i^j\right) + R_i \stackrel{(d)}{=} \E_{\x,y}\left[\hat{\sigma}_l(\f_i^T \x) y \right] + R_i,
\end{align}
where $R_i := \sum_{j=l}^{\infty} \frac{1}{j!} \mu_j \Tilde{\mu}_j \eta_i^j$ to reach (c). To reach (a)-(b), we use the orthogonality of Hermite polynomials (Lemma \ref{lemma:hermite_correlated}). Finally, we get (d) by the definition of $\hat{\sigma}_l$. For $|R_i|$, we derive an upper bound as follows
\begin{align}
    |R_i| &= \left|\sum_{j=l}^{\infty} \frac{1}{j!} \mu_j \Tilde{\mu}_j \eta_i^j\right| \stackrel{(a)}{\leq} \sum_{j=l}^{\infty} \frac{1}{j!} |\mu_j \Tilde{\mu}_j| |\eta_i^j| \stackrel{(b)}{\leq} \frac{C}{k},
\end{align}
for some $C > 0$. The triangle inequality is used to reach (a) while we get (b) using $|\eta_i^l| = \mathcal{O}(1/k)$ (by the assumption of the lemma). Finally, we have
\begin{align*}
    \left\|\E_{(\x,y)} \left[\sigma(\F\x) y \right] - \E_{(\x,y)} \left[\hat{\sigma}_l(\F\x) y \right]\right\| &\leq \sqrt{\sum_{i=1}^{k} \left|\E[\sigma(\f_i^T \x) y] - \E[\hat{\sigma}_l(\f_i^T \x) y] \right|^2}= \sqrt{\sum_{i=1}^{k} R_i^2} \leq \frac{C}{\sqrt{k}},
\end{align*}
which completes the proof.
\end{proof}
\end{lemma}

\bibliography{iclr2025_conference}
\bibliographystyle{iclr2025_conference}

\newpage
\setcounter{section}{0}
\setcounter{subsection}{0}

\renewcommand\thesection{\arabic{section}}
\renewcommand\thesubsection{\thesection\arabic{subsection}}
\renewcommand\thesubsubsection{\thesubsection\arabic{subsubsection}}

\renewcommand\thesection{SM\arabic{section}}
\renewcommand\thesubsection{\thesection\arabic{subsection}}
\renewcommand\thesubsubsection{\thesubsection\arabic{subsubsection}}

\renewcommand\theequation{SM\arabic{equation}}
\renewcommand\thefigure{SM\arabic{figure}}
\renewcommand\thetable{SM\arabic{table}}
\setcounter{equation}{0}
\setcounter{figure}{0}
\setcounter{table}{0}

\setcounter{theorem}{0}

\renewcommand\thetheorem{SM\arabic{theorem}}

\section*{SUPPLEMENTARY MATERIALS}

\section{Auxiliary Results for the Proofs}
\label{supplement:auxiliary}
In this section, we provide auxiliary results that are used in the proofs given in the previous sections.

\begin{lemma}
    Suppose assumptions (S.1)-(S.6) in Section \ref{sec:assumptions} and additional assumptions S.7-S.8 in Appendix \ref{appendix:equivalence_proof}. Then, 
    \begin{align}
        \G_\sigma \stackrel{\Prob}{\to} e_{\G} \quad \textit{ if and only if } \quad \G_{\hat{\sigma}} \stackrel{\Prob}{\to} e_{\G}, \label{supplement_eq:generalization_equivalence_lemma}
    \end{align}
    if the following holds,
    \begin{align}
        \frac{\Phi_{\A}(\tau_1, \tau_2) }{k} \stackrel{\Prob}{\to} \phi \quad \text{if and only if} \quad \frac{\Phi_{\B}(\tau_1, \tau_2) }{k} \stackrel{\Prob}{\to} \phi,
    \end{align}
where $\Phi_{\bR}(\tau_1, \tau_2)$ is the perturbed problem defined in \eqref{eq:perturbed_objective}. Also, $\A := [\a_1, \a_2, \dots, \a_m]^T$ and $\B := [\b_1, \b_2, \dots, \b_m]^T$ where $\a_i :=\sigma(\F \x_i)$ and $\b_i := \hat{\sigma}(\F \x_i)$ for all $i \in \{1,2,\dots,m\}$. 
\begin{proof}
We start by writing $\G_\sigma$ as follows
\begin{align}
    \G_\sigma :&= \E_{(\x,y)} (y - \hat{\w}_\sigma^T\sigma(\F \x))^2, \\
    &= \E [y^2] - 2\hat{\w}_\sigma^T \E[\sigma(\F \x)y] + \hat{\w}_\sigma^T \E[\sigma(\F \x) \sigma(\F \x)^T]\hat{\w}_\sigma,\\
    &= \E[y^2] - 2\hat{\w}_\sigma^T \Sigma_{xy} + \hat{\w}_\sigma^T \Sigma_{x} \hat{\w}_\sigma,\\
    &= \E[y^2] - 2\pi_{\A} + \rho_{\A},
    \label{supplement_eq:generalization_in_terms_of_rho_pi}
\end{align}
where we define $\pi_{\A} := \hat{\w}_0^T \Sigma_{xy}$ and $\rho_{\A} := \hat{\w}_0^T \Sigma_{x} \hat{\w}_0$ in the last step. Note that $\hat{\w}_\sigma = \hat{\w}_0$ and $\hat{\w}_{\hat{\sigma}} = \hat{\w}_{m}$ by definition \eqref{eq:w_t}. Similarly, we can arrive at $\G_{\hat{\sigma}} = \E[y^2] - 2 \pi_{\B} + \rho_{\B} + R$ for activation function $\hat{\sigma}$ where $\pi_{\B} := \hat{\w}_{m}^T \Sigma_{xy}$ and $\rho_{\B} := \hat{\w}_{m}^T \Sigma_{x} \hat{\w}_{m}$. Furthermore, we define $R$ as the remainder term as follows 
\begin{align}
    R &:= \hat{\w}_{m}^T \left(\E[\hat{\sigma}(\F\x) \hat{\sigma}(\F\x)^T] - \Sigma_x\right) \hat{\w}_{m} - 2\hat{\w}_{m}^T \left(\E[\hat{\sigma}(\F\x) y] - \Sigma_{xy}\right),
\end{align}
which can be bounded as
\begin{align}
    \mathbb{E}_{\backslash \boldsymbol{F}} |R| &\leq \mathbb{E}_{\backslash \boldsymbol{F}}\left(\left\|\hat{\w}_{m}\right\|^2\right) \left\|\E[\hat{\sigma}(\F\x) \hat{\sigma}(\F\x)^T] - \Sigma_x\right\|  + 2\mathbb{E}_{\backslash \boldsymbol{F}}\left(\left\|\hat{\w}_{m}\right\|\right) \left\|\E[\hat{\sigma}(\F\x) y] - \Sigma_{xy}\right\|,\\
    &= o(1),
\end{align}
where we use a bound on $\mathbb{E}_{\backslash \boldsymbol{F}}[\left\|\hat{\w}_{m}\right\|^l]$ (Lemma \ref{supplement_lemma:w_t_bound}) with the equivalence of covariance matrices \eqref{eq:Sigma_x} and the equivalence of cross-covariance vectors \eqref{eq:Sigma_xy} to reach the last line. Now, to prove \eqref{supplement_eq:generalization_equivalence_lemma},  we need to show $\rho_{\A} \stackrel{\Prob}{\to} \rho^* = \frac{\partial}{\partial \tau_1} q^*(0,0)$ while $\rho_{\B} \stackrel{\Prob}{\to} \rho^*$  and $\pi_{\A} \stackrel{\Prob}{\to} \pi^* = \frac{\partial}{\partial \tau_2} q^*(0,0)$ while $\pi_{\B} \stackrel{\Prob}{\to} \pi^*$ under assumption S.8. We show $\rho_{\A} \stackrel{\Prob}{\to} \rho^*$ in the following while the rest can be shown analogously. First, observe that, for any $\tau_1, \tau_2 \in (-\tau_*, \tau_*)$,
\begin{equation}
    \Phi_{\A}(\tau_1, \tau_2) \leq \Phi_{\A}(0, 0) + \tau_1 k \hat{\w}_0^T \Sigma_{x} \hat{\w}_0 + \tau_2 k \hat{\w}_0^T \Sigma_{xy},
\end{equation}
which we arrive at by using the definition of the perturbed optimization \eqref{eq:perturbed_objective}. Then, for any $\tau \in (0, \tau_*)$,
\begin{equation}
    \frac{\Phi_{\A}(\tau, 0) - \Phi_{\A}(0, 0)}{k \tau} \leq \rho_{\A} \leq \frac{\Phi_{\A}(-\tau, 0) - \Phi_{\A}(0, 0)}{-k \tau}.
    \label{supplement_eq:perturbed_inequality}
\end{equation}
For a fixed $\epsilon > 0$, there exists $\delta > 0$ such that 
\begin{equation}
    \left| \frac{q^*(\delta,0) - q^*(0,0)}{\delta} - \rho^* \right| \leq \epsilon/3,
    \label{supplement_eq:q_derivative}
\end{equation}
due to assumption S.8 in Appendix \ref{appendix:equivalence_proof}. Focusing on the inequality on the left in \eqref{supplement_eq:perturbed_inequality}, we get:
\begin{align}
    \Prob(\rho_{\A} - \rho^* < -\epsilon) &\leq \Prob\left(\frac{\Phi_{\A}(\delta, 0) - \Phi_{\A}(0, 0)}{k \delta} - \rho^* < -\epsilon\right),\\
    &\leq  \Prob( | \Phi_{\A}(\delta, 0)/k - q^*(\delta,0) | > \delta \epsilon/3) +  \Prob( | \Phi_{\A}(0, 0)/k - q^*(0,0) | > \delta \epsilon/3),\\
    &= 0,
\end{align}
where we use \eqref{supplement_eq:q_derivative} in the middle line and $\Phi_{\A}(\tau_1, \tau_2)/k \stackrel{\Prob}{\to} q^*(\tau_1,\tau_2)$ is used in the last line. 
Applying the same steps to the inequality on the right in \eqref{supplement_eq:perturbed_inequality}, we arrive at $\Prob(\rho_{\A} - \rho^* > \epsilon) = 0$. Therefore, we conclude that $\rho_{\A} \stackrel{\Prob}{\to} \rho^*$. Using the same reasoning that we used to show $\rho_{\A} \stackrel{\Prob}{\to} \rho^*$, we may also show $\rho_{\B} \stackrel{\Prob}{\to} \rho^*$, $\pi_{\A} \stackrel{\Prob}{\to} \pi^*$, and $\pi_{\B} \stackrel{\Prob}{\to} \pi^*$ but we omit them here. Since we have $\rho_{\A} \stackrel{\Prob}{\to} \rho^*$, $\rho_{\B} \stackrel{\Prob}{\to} \rho^*$, $\pi_{\A} \stackrel{\Prob}{\to} \pi^*$, and $\pi_{\B} \stackrel{\Prob}{\to} \pi^*$, we reach \eqref{supplement_eq:generalization_equivalence_lemma} by using \eqref{supplement_eq:generalization_in_terms_of_rho_pi}.
\end{proof}
\label{supplement_lemma:perturbed_equivalence}
\end{lemma}

\begin{lemma}
Suppose the setting in Appendix \ref{appendix:equivalence_proof}. Then, for any $\phi \in \R$,
    \begin{align}
    \frac{\Phi_{\A}(\tau_1, \tau_2) }{k} \stackrel{\Prob}{\to} \phi \quad \text{if and only if} \quad \frac{\Phi_{\B}(\tau_1, \tau_2) }{k} \stackrel{\Prob}{\to} \phi, \label{supplement_eq:probabilistic_convergence_lemma}
\end{align}
if the following holds:
\begin{align}
    \left| \E \varphi \left( \frac{1}{k} \Phi_\A \right) - \E \varphi \left( \frac{1}{k} \Phi_\B \right) \right| \leq \max \left\{||\varphi||_\infty, ||\varphi^\prime||_\infty, ||\varphi^{\prime\prime}||_\infty \right\} \kappa_k,
    \label{supplement_eq:test_bound_lemma}
\end{align}
for some $\kappa_k = o(1)$ and every bounded test function $\varphi(x)$ which has bounded first two derivatives.
\begin{proof}
    The proof is adapted from Section II-D in \cite{hu2022universality}. We need to select an appropriate test function to start the proof. For any fixed $\epsilon > 0$ and $\phi$, let 
    \begin{equation}
        \varphi_\epsilon(x) := \mathds{1}_{|x| \geq 3\epsilon/2} * \zeta_{\epsilon/2}(x-\phi),
    \end{equation}
    where $*$ denotes the convolution operation and $\zeta_{\epsilon}(x) := \epsilon^{-1}\zeta(x/\epsilon)$ is a scaled mollifier. We define a standard mollifier as:
    \begin{equation}
       \zeta(x) := \begin{cases}
    ce^{-1/(1-x^2)}, & \text{if } |x| < 1,\\
    0,              & \text{otherwise,}
\end{cases}
    \end{equation}
    for some constant $c$ ensuring $\int_{\R} \zeta(x) dx = 1$. We may easily verify that $||\varphi^{\prime}_\epsilon(x)||_\infty < C/\epsilon$ and $||\varphi^{\prime\prime}_\epsilon(x)||_\infty < C/\epsilon^2$ for some constant $C$. Furthermore, 
    \begin{equation}
        \mathds{1}_{|x-\phi| \geq 2\epsilon} \leq  \varphi_\epsilon(x) \leq \mathds{1}_{|x-\phi| \geq \epsilon}.
        \label{supplement_eq:mollifier_inequality}
    \end{equation}
    Let $x = \Phi_\A/k$ in \eqref{supplement_eq:mollifier_inequality} and take the expectation:
    \begin{equation}
        \Prob(|\Phi_\A/k - \phi|\geq 2\epsilon) \leq \E[\varphi_\epsilon(\Phi_\A/k)].
    \end{equation}
    Similarly, let $x = \Phi_\B/k$ in \eqref{supplement_eq:mollifier_inequality} and take the expectation:
    \begin{equation}
        \E[\varphi_\epsilon(\Phi_\B/k)] \leq \Prob(|\Phi_\B/k - \phi|\geq \epsilon).
    \end{equation}
    Using \eqref{supplement_eq:test_bound_lemma} together with the last two equations, we arrive at
    \begin{equation}
        \Prob(|\Phi_\A/k - \phi|\geq 2\epsilon) \leq \Prob(|\Phi_\B/k - \phi|\geq \epsilon) + \text{max}\left\{C, C/\epsilon, C/\epsilon^2 \right\} \kappa_k,
    \end{equation}
    which leads to
        \begin{equation}
        \Prob(|\Phi_\A/k - \phi|\geq 2\epsilon) \leq \Prob(|\Phi_\B/k - \phi|\geq \epsilon) + \frac{C \kappa_k}{\epsilon^2},
    \end{equation}
    for $\epsilon \in (0,1)$ satisfying $\kappa_k/\epsilon^2 = o(1)$. Applying the same steps after switching $\A$ with $\B$, we get
    \begin{equation}
        \Prob(|\Phi_\B/k - \phi|\geq 2\epsilon) \leq \Prob(|\Phi_\A/k - \phi|\geq \epsilon) + \frac{C \kappa_k}{\epsilon^2}.
    \end{equation}
    Combining the last two results and letting $\epsilon \to 0^+$ with $\kappa_k/\epsilon^2 = o(1)$, we reach \eqref{supplement_eq:probabilistic_convergence_lemma}.
\end{proof}
\label{supplement_lemma:test_function_equivalence}
\end{lemma}

\begin{lemma}
    Suppose the definitions in Appendix \ref{appendix:equivalence_proof}. Then, the following holds:
     \begin{equation}
         \Phi_{t-1} = \Psi_t(\a_t), \qquad \textit{and} \qquad \Phi_{t} = \Psi_t(\b_t).
     \end{equation}
     \begin{proof}
         We apply Taylor expansion to $\Phi_{t-1}$ around $\Phi_{\backslash t}$:
         \begin{align}
             \Phi_{t-1} = \Phi_{\backslash t} + \min _{\w \in \mathbb{R}^k} \left\{ (\w - \hat{\w}_{\backslash t})^T\boldsymbol{d}_{\backslash t} + \frac{1}{2}\left(\w-\hat{\w}_{\backslash t}\right)^T \boldsymbol{H}_{\backslash t}\left(\w-\hat{\w}_{\backslash t}\right) + \left(\w^T \a_t - y_t \right)^2 \right\},
         \end{align}
         where $\boldsymbol{d}_{\backslash t}$ and $\boldsymbol{H}_{\backslash t}$ are respectively gradient vector and Hession matrix of $\Phi_{\backslash t}$ evaluated at $\hat{\w}_{\backslash t}$. Note that we do not need higher order terms since $\Phi_{\backslash t}$ only involves quadratic terms, which means higher order terms are equal to $0$. Furthermore, the gradient vector $\boldsymbol{d}_{\backslash t}$ is equal to $0$ due to the first-order optimality condition in \eqref{eq:hat_w}. This leads us to:
         \begin{align}
             \Phi_{t-1} = \Phi_{\backslash t} + \min _{\w \in \mathbb{R}^k} \left\{\frac{1}{2}\left(\w-\hat{\w}_{\backslash t}\right)^T \boldsymbol{H}_{\backslash t}\left(\w-\hat{\w}_{\backslash t}\right) + \left(\w^T \a_t - y_t \right)^2 \right\} = \Psi_t(\a_t).
         \end{align}
         Similarly, one can show $\Phi_{t} = \Psi_t(\b_t)$ using the same reasoning, which is omitted here.
     \end{proof}
    \label{supplement_lemma:equality_of_surrogate}
\end{lemma}

\begin{lemma}
    Suppose the definitions in Appendix \ref{appendix:equivalence_proof}. For some $\epsilon$, the following holds:
    \begin{equation}
        \max \left\{\mathbb{E}_{\backslash \boldsymbol{F}}\left(\Psi_t\left(\boldsymbol{b}_t\right)-\Phi_{\backslash t}\right)^2, \mathbb{E}_{\backslash \boldsymbol{F}}\left(\Psi_t\left(\boldsymbol{a}_t\right)-\Phi_{\backslash t}\right)^2\right\} \leq k^{1-\epsilon}\text{polylog }k.
    \end{equation}
    \begin{proof}
        Let $\br_t = \a_t$ or $\br_t = \b_t$. Then, 
        \begin{align}
            \mathbb{E}_{\backslash \boldsymbol{F}}\left(\Psi_t\left(\br_t\right)-\Phi_{\backslash t}\right)^2 &\leq \mathbb{E}_{\backslash \boldsymbol{F}}\left(\hat{\w}_{\backslash t}^T \br_t - y_t \right)^4,\\
            &\leq \sqrt{\mathbb{E}_{\backslash \boldsymbol{F}} \left( \left| \hat{\w}_{\backslash t}^T \br_t \right| +1 \right)^{8}} \text{ polylog } k,\\
            &\leq Q \left(\|\hat{\w}_{\backslash t}\| \right) \|\hat{\F}\|^4  \text{ polylog } k,\\
            &\leq k^{1-\epsilon} \text{ polylog } k
        \end{align}
        where $Q(x)$ is some finite-degree polynomial, $\hat{\F} := \F (\I_n + (\sqrt{1+\theta} - 1) \bgamma\bgamma^T)$ and some $\epsilon > 0$ satisfying $\theta^2 \leq n^{1-\epsilon}$. We use \eqref{eq:surrogate_vs_loss} in the first step. Then, we reach the second line using Lemma \ref{supplement_lemma:loss_power_bound}. In the third step, we use Lemma \ref{supplement_lemma:wTa_bound}. Finally, we use \eqref{eq:f_cond2} and Lemma \ref{lemma:w_t_bound} to conclude the proof. 
    \end{proof}
\label{supplement_lemma:surrogate_vs_leave_one_out}
\end{lemma}

\begin{lemma}
    Suppose the definitions in Appendix \ref{appendix:equivalence_proof}. Then, the following holds:
    \begin{align}
        \mathbb{E}_{\backslash \boldsymbol{F}}\left(\hat{\w}_{\backslash t}^T \br - y_t \right)^{2l} \leq \sqrt{\mathbb{E}_{\backslash \boldsymbol{F}} \left( \left| \hat{\w}_{\backslash t}^T \br \right| +1 \right)^{4l}} \text{ polylog } k,
        \label{supplement_eq:loss_power_bound}
    \end{align}
    for $l \in \mathbb{Z}^+$.
    \begin{proof}
        Let $y_t := \sigma_*(s_t)$ where $s_t := \frac{\bxi^T \x_t}{\sqrt{1+\theta\alpha^2}}$. Then, for any $x \in \R$,
        \begin{align}
           (x-y_t)^2 &\leq |x|^2 + 2|x||y_t| + |y_t|^2,\\
           &\leq |x|^2 +2|x| C \left(| s_t|^K + 1 \right) + C^2 \left(| s_t|^K + 1 \right)^2,\\
           &\leq \hat{C}  (|x| + 1)^2  \left(| s_t|^K + 1 \right)^2,
        \end{align}
        for some $C, \hat{C} >0$. To reach the second line, we use $|y_t| \leq C \left(| s_t|^K + 1 \right)$ due to assumption (S.3). Using this result, we get:
        \begin{align}
        \mathbb{E}_{\backslash \boldsymbol{F}}\left(\hat{\w}_{\backslash t}^T \br - y_t \right)^{2l} &\leq \hat{C}^l \mathbb{E}_{\backslash \boldsymbol{F}} \left[ \left( \left| \hat{\w}_{\backslash t}^T \br \right| +1 \right)^{2l} \left( \left| s_t \right|^K + 1 \right)^{2l} \right],\\ 
        &\leq \hat{C}^l \sqrt{\mathbb{E}_{\backslash \boldsymbol{F}} \left( \left| \hat{\w}_{\backslash t}^T \br \right| +1 \right)^{4l}} \sqrt{\mathbb{E}_{\backslash \boldsymbol{F}} \left( \left| s_t \right|^K + 1 \right)^{4l}}.
    \end{align}
   $s_t \sim \Normal(0,1)$ so $\Prob(|s_t| > \epsilon) \leq 2 e^{-\epsilon^2/2}$ \cite[Eq. (2.10)]{vershynin2018high}. Therefore, we can reach \eqref{supplement_eq:loss_power_bound} by using moment bounds for $|s_t|$ from Lemma \ref{supplement_lemma:concentration_lipschitz_functions}.
    \end{proof}
    \label{supplement_lemma:loss_power_bound}
\end{lemma}

\begin{lemma}
Suppose assumptions (S.1)-(S.6). Let $\a := \sigma(\F \x)$. Furthermore, let $\hat{\F} := \F (\I_n + (\sqrt{1+\theta} - 1) \bgamma\bgamma^T)$. Then, there exist some $c, C > 0$ such that the  following holds
\begin{align}
     \Prob_{\backslash \F} \left( \left| \w^T \a \right| \geq \epsilon \right) &\leq 2 e^{\frac{-\epsilon^2}{c \|\w\|^2 \|\hat{\F}\|^2 \text{ polylog } k}}, \label{supplement_eq:concentration_wTa}\\
      \mathbb{E}_{\backslash \F} \left( \left| \w^T \a \right|^l \right) &\leq (l!) \left( C\|\w\|^2 \|\hat{\F}\|^2 \text{ polylog } k \right)^{l/2},\label{supplement_eq:concentration_wTa_power}
\end{align}
for any $l \in \mathbb{Z}^+$, any fixed $\w \in \R^k$ and $\epsilon \geq 0$.
\begin{proof}
    Let $\g \sim \Normal(0, \I_k)$. One can verify that the function $f(\g) := \w^T \a = \w^T \sigma(\hat{\F} \g)$ is $(\|\sigma^{\prime}\|_\infty \|\w\| \|\hat{\F}\|)$-Lipschitz continuous. Furthermore, $\E_{\backslash \F} \left( \w^T \a \right) = 0$ due to $\sigma(x)$ being an odd function. Furthermore, $\sigma^{\prime}(z) \leq \text{ polylog } k$ with high probability for $z \sim \Normal(0,\hat{C})$ with all $\hat{C} > 0$, which allows us to continue with $\|\sigma^{\prime}\|_\infty \leq \text{ polylog } k$ by using a truncation argument (similar to Lemma \ref{lemma:truncated_expectation}). Therefore, we can reach \eqref{supplement_eq:concentration_wTa}-\eqref{supplement_eq:concentration_wTa_power} using Lemma \ref{supplement_lemma:concentration_lipschitz_functions}.
\end{proof}
\label{supplement_lemma:wTa_bound}
\end{lemma}

\begin{lemma}[Concentration of Lipschitz Functions - {\cite[Theorem 3]{hu2022universality}}]
Consider $\g \sim \Normal(0,\I_k)$. Then, for any $\kappa$-Lipschitz function $f:\R^k \to \R$ and $\epsilon \geq 0$,
\begin{equation}
    \Prob(|f(\g) - \E[f(\g)]| \geq \epsilon) \leq 2 e^{-\epsilon^2/(4\kappa^2)}.
\end{equation}
Furthermore, let $x$ be a random variable. If $x$ satisfying $\Prob(|x| > v) \leq c e^{-C v}$ for some $C, c > 0$,
\begin{equation}
    \E\left[ |x|^l \right] \leq c l C^{-l} \int_{0}^{\infty} e^{-v} v^{l-1} dv = c(l!) C^{-l} \quad \text{for any $l \in \mathbb{Z}^+$}.
\end{equation}
Similarly, if $x$ satisfying $\Prob(|x| > v) \leq c e^{-C v^2}$ for some $C, c > 0$,
\begin{equation}
    \E\left[ |x|^l \right] \leq  2c(l!) C^{-l/2} \quad \text{for any $l \in \mathbb{Z}^+$}.
\end{equation}
\begin{proof}
    See \cite[Theorem 1.3.4]{talagrand2010mean}.
\end{proof}
\label{supplement_lemma:concentration_lipschitz_functions}    
\end{lemma}

\begin{lemma}
    Suppose the definitions in Appendix \ref{appendix:equivalence_proof}. Let $\epsilon >0$ be a constant satisfying $\theta \leq n^{1/2-\epsilon/2}$. For $\br_t = \a_t$ or $\br_t = \b_t$, we have
    \begin{align}
        \Prob(|  \nu_t(\br_t) - \E_t [\nu_t(\br_t)] | > \delta) \leq c e^{-Ck^{\epsilon} \delta^2 / \text{polylog } k}, \label{supplement_eq:nu_t_tail_bound}
    \end{align}
     for some $c,C >0$. Furthermore,
    \begin{align}
        \E_t[| \nu_t(\br_t) - \E_t [\nu_t(\br_t)]|^l] \leq 2 c (l!) (C k^{\epsilon} / \text{polylog } k)^{-l/2}, \label{supplement_eq:nu_t_moment_bound}
    \end{align}
    for some $c,C >0$ and any $l \in \mathbb{Z}^+$. Finally,
    \begin{align}
        \left|\E_t \nu_t(\a_t) - \E_t [\nu_t(\b_t)]\right| = \mathcal{O}(k^{-\varsigma}), \label{supplement_eq:nu_a_t_vs_nu_b_t}
    \end{align}
    for some $\varsigma >0$.
    \begin{proof}
    Consider
    \begin{align}
        \nu_t(\br_t) &= \br_t^{T} \boldsymbol{H}_{\backslash t}^{-1} \br_t = \frac{\br_t^{T} \bar{\boldsymbol{H}}_{\backslash t}^{-1} \br_t}{k},
    \end{align}
    where $ \bar{\boldsymbol{H}}_{\backslash t}$ is defined as follows to take the $1/k$ scaling out
    \begin{align}
\bar{\boldsymbol{H}}_{\backslash t} := &  \frac{2}{k}\sum_{i=1}^{t-1} \b_i \b_i^T + \frac{2}{k} \sum_{i=t+1}^m \a_i \a_i^T + \frac{1}{k} \nabla^2 Q\left(\hat{\w}_{\backslash t}\right).
\end{align}
Then, using $s$-strong convexity of $Q(\w)/m$ due to assumption S.7, we have $\bar{\boldsymbol{H}}_{\backslash t} \succeq (m/k) s \I_k$, which lead to $\|\bar{\boldsymbol{H}}_{\backslash t}^{-1}\| \leq (k/m)(1/s)$. Furthermore, we can rewrite $\a_i = \sigma(\F \x_i)$ equivalently as $\sigma(\hat{\F} \g)$ for $\hat{\F} := \F (\I_n + (\sqrt{1+\theta} - 1)\bgamma\bgamma^T)$ and $\g \sim \Normal(0,\I_n)$. We have $\|\hat{\F}\| \leq k^{1/4 - \epsilon/4} \text{ polylog } k$ (for some $\epsilon >0$ satisfying $\theta \leq n^{1/2-\epsilon/2}$), which is due to \eqref{eq:f_cond2}. Furthermore, we have $\|\sigma^{\prime}\|_\infty \leq \text{ polylog } k$ by using a truncation argument (similar to Lemma \ref{lemma:truncated_expectation}) as mentioned in the proof of Lemma \ref{supplement_lemma:wTa_bound}. Then, we use \cite[Lemma 1]{louart2018random} with the bounds for $\|\hat{\F}\|$ and $\|\bar{\boldsymbol{H}}_{\backslash t}^{-1}\|$ to reach \eqref{supplement_eq:nu_t_tail_bound}. Finally, \eqref{supplement_eq:nu_t_moment_bound} can be obtained using Lemma \ref{supplement_lemma:concentration_lipschitz_functions}. To show \eqref{supplement_eq:nu_a_t_vs_nu_b_t}, we proceed as
\begin{align}
    \left|\E_t \nu_t(\a_t) - \E_t [\nu_t(\b_t)]\right| &= \frac{1}{k} \left| \E_t \left[\a_t^{T} \bar{\boldsymbol{H}}_{\backslash t}^{-1} \a_t\right] - \E_t \left[\b_t^{T} \bar{\boldsymbol{H}}_{\backslash t}^{-1} \b_t \right]  \right|,\\
    &=\frac{1}{k} \left|\text{Tr}\left(\bar{\boldsymbol{H}}_{\backslash t}^{-1} \left(\E_t\left[\a_t\a_t^T\right] - \E_t\left[\b_t\b_t^T\right]\right)\right) \right|,\\
    &=\mathcal{O}(k^{-\varsigma}),
\end{align}
for some $\varsigma >0$, where $\text{Tr}$ denotes the trace while we use $\|\bar{\boldsymbol{H}}_{\backslash t}^{-1}\| \leq (k/m)(1/s)$ and \eqref{eq:Sigma_x} to reach the last line.
    \end{proof}
    \label{supplement_lemma:nu_tail_bound}
\end{lemma}

\begin{lemma}
    Suppose the definitions in Appendix \ref{appendix:equivalence_proof}. Then, the following holds:
    \begin{align}
        \mathbb{E}_{\backslash \F}  \|\hat{\w}_t\|^l &\leq \text{polylog } k, \label{eq:hat_w_bound}\\ 
        \mathbb{E}_{\backslash \F}  \|\hat{\w}_{\backslash t}\|^l &\leq \text{polylog } k, \label{eq:hat_w_backslash_t_bound}
    \end{align}
    for any $l \in \mathbb{Z}^+$ and any $t \in \{0,\dots,m\}$.
    \begin{proof}
        We show \eqref{eq:hat_w_bound} here while \eqref{eq:hat_w_backslash_t_bound} can be proved similarly. Let $\br_{i} := \b_i$ for $i \in \{1,\dots,k\}$ and $\br_{i} := \a_i$ for $t \in \{k+1,\dots,m\}$. Furthermore, let $Q(\w) := m\lambda||\w||_2^2 + \tau_1 k \w^T \Sigma_x \w + \tau_2 k \w^T \Sigma_{xy}$. Then, 
        \begin{align}
            \frac{1}{m}Q(\hat{\w}_t) &\leq \frac{1}{m} \sum_{i=1}^{m} \left(\hat{\w}_t^T\br_i) - y_i \right)^2 + \frac{1}{m} Q(\hat{\w}_t) \leq \frac{1}{m}\sum_{i=1}^{m} y_i^2 + \frac{1}{m}Q(\boldsymbol{0}),
        \end{align}
        since $\hat{\w}_t$ is the optimal solution defined in \eqref{eq:w_t}. Furthermore, $Q(\w)/m$ is $s$-strongly convex by assumption S.7 in Appendix \ref{appendix:equivalence_proof}. Using this fact, we reach
        \begin{align}
            \frac{1}{m} Q(\hat{\w}_t) &\geq \frac{1}{m}Q(\boldsymbol{0}) + \frac{1}{m} \nabla Q(\boldsymbol{0})^T \hat{\w}_t + (s/2) \|\hat{\w}_t\|^2,\\
            &\geq \frac{1}{m}Q(\boldsymbol{0}) - \frac{1}{m}\|\nabla Q(\boldsymbol{0})\| \|\hat{\w}_t \| + (s/2) \|\hat{\w}_t\|^2.
        \end{align}
        We may then combine the found lower and upper bounds for $Q(\hat{\w}_t)/m$ as follows:
        \begin{equation}
            (s/2) \|\hat{\w}_t\|^2 - \frac{1}{m}\|\nabla Q(\boldsymbol{0})\| \|\hat{\w}_t \| \leq \frac{1}{m}\sum_{i=1}^{m} y_i^2,
        \end{equation}
        which leads to 
        \begin{align}
            \|\hat{\w}_t\| &\leq \frac{\|\nabla Q(\boldsymbol{0})\|/m + \sqrt{\|\nabla Q(\boldsymbol{0})\|^2/m^2 + 2(s/m) \sum_{i=1}^{m} y_i^2}}{s},\\
            &\leq \frac{2\|\nabla Q(\boldsymbol{0})\|/m + \sqrt{2(s/m) \sum_{i=1}^{m} y_i^2}}{s},\\
            &\leq \left(\frac{1}{m} \sum_{i=1}^{m} y_i^2\right)^{1/2} \text{ polylog } k,
        \end{align}
        where we use $\|\nabla Q(\boldsymbol{0})\|/m = |\tau_2| (k/m) \|\Sigma_{xy}\| \leq \text{polylog } k$ since we have $(k/m) \in (0,\infty)$ by assumption (S.4) in Section \ref{sec:assumptions}, $\tau_2 = \mathcal{O}(1/\sqrt{k})$ by assumption S.7 in Appendix \ref{appendix:equivalence_proof} and one can easily show $\|\Sigma_{xy}\| \leq \sqrt{k} \text{ polylog } k$ using \eqref{eq:Sigma_xy_decomposition}. Then, we consider $l$-th power of $\|\hat{\w}_t\|$ as
        \begin{equation}
           \|\hat{\w}_t\|^l \leq \left(\frac{1}{m} \sum_{i=1}^{m} y_i^2\right)^{l/2} \text{ polylog } k.
        \end{equation}
        Let $y_i := \sigma_*(s_i)$ where $s_i := \frac{\bxi^T \x_i}{\sqrt{1+\theta\alpha^2}}$. Using $|y_i| \leq C \left(| s_i|^K + 1 \right)$ due to assumption (S.3), we get
        \begin{align}
            \mathbb{E}_{\backslash \F}\|\hat{\w}_t\|^l &\leq  \mathbb{E}_{\backslash \F} \left(\frac{1}{m} \sum_{i=1}^{m}  \left(| s_i|^K + 1 \right)^2\right)^{l/2} \text{ polylog } k.
        \end{align}
        Finally, $s_i \sim \Normal(0,1)$ so we reach \eqref{eq:hat_w_bound} by using moment bounds for $|s_i|$ from Lemma \ref{supplement_lemma:concentration_lipschitz_functions}. Similarly, \eqref{eq:hat_w_backslash_t_bound} can shown using analogous steps.
    \end{proof}
    \label{supplement_lemma:w_t_bound}
\end{lemma}

\begin{lemma}
    Let $f: \R \to \R$ be a function satisfying $|f(x)| \leq C (1 + |x|^{K})$ for all $x \in \R$ and some constants $C > 0, K \in \mathbb{Z}^+$. Furthermore, define a truncated version of $f$ as
    \begin{align}
        f_{\epsilon}(x) := \begin{cases}
        f(x), & \text{if } |x| < \epsilon,\\
        0, & \text{otherwise,}
    \end{cases}
    \end{align}
    for $\epsilon > 0$. If $\epsilon =2log k$, the following holds for any constant $a > 0$ and $z \sim \Normal(0,1)$, 
    \begin{equation}
        \mathbb{E}_z|f(a z) - f_{\epsilon}(a z) | = \frac{\text{ polylog } k}{k^{log k}}.
    \end{equation}
    \begin{proof}
        We start by using $1 = \mathds{1}_{|z|<\epsilon} + \mathds{1}_{|z|\geq \epsilon}$ as follows 
        \begin{align}
            \mathbb{E}_z|f(a z) - f_{\epsilon}(a z) | &\leq \mathbb{E}_z [| f(a z) -f_{\epsilon}(a z)| (\mathds{1}_{|z|<\epsilon} + \mathds{1}_{|z|\geq \epsilon})],\\
            &= \mathbb{E}_z [| f(a z) - f_{\epsilon}(a z) | \mathds{1}_{|z|\geq \epsilon} ],\\
            &= \mathbb{E}_z [| f(a z) | \mathds{1}_{|z|\geq \epsilon} ]\\
            &\stackrel{(a)}{\leq} \sqrt{\mathbb{E}_z [ f(a z)^2]} \sqrt{\mathbb{E}_z [(\mathds{1}_{|z|\geq \epsilon})^2 ]},\\
            &= \sqrt{\mathbb{E}_z [ f(a z)^2]} \sqrt{\Prob(|z| \geq \epsilon)},\\
            &\stackrel{(b)}{\leq} C\sqrt{\mathbb{E}_z [ (a^{2K}z^{2K} + 2a^K|z|^K + 1)]} \sqrt{2e^{-\epsilon^2/2}},\\
            &\stackrel{(c)}{\leq} \frac{\text{ polylog } k}{e^{log^2(k)}} = \frac{\text{ polylog } k}{k^{log k}},
        \end{align}
        where we use Cauchy–Schwarz inequality to reach (a) while (b) is due to $|f(x)| \leq C (1 + |x|^{K})$ and $\Prob(|z| > \epsilon) \leq 2 e^{-\epsilon^2/2}$ \cite[Eq. (2.10)]{vershynin2018high}. Finally, we reach (c) by using moment bounds for $|z|$ from Lemma \ref{supplement_lemma:concentration_lipschitz_functions}.
    \end{proof}
    \label{supplement_lemma:truncated_expectation}
\end{lemma}

\begin{lemma}
    Let $H_i(x)$ be the probabilist's $i$-th Hermite polynomial and $z_1, z_2 \stackrel{i.i.d}{\sim} \Normal(0,1)$. Then, for any $\rho \in [0,1]$ and any $c \geq |\rho|$,
    \begin{align}
        \mathbb{E}_{z_1, z_2}[H_i(\rho z_1 + \sqrt{c^2- \rho^2} z_2) H_j(z_1)] = (i!) \rho^i \delta_{i,j},
    \end{align}
    where $\delta_{i j}$ is Kronecker delta function.
    \begin{proof}
        Let $a= (\rho/c) z_1 +  \sqrt{1- \rho^2/c^2} z_2$ and $b = z_1$. Observe that $(a,b)$ is jointly Gaussian with $\E[a^2] = \E[b^2] = 1$ and $\E[ab] = (\rho/c)$. Furthermore, let $p(a, b)$ be the joint probability density function of $(a,b)$ while $p(a)$ and $p(b)$ denoting the individual probability density functions. Using Mehler's formula \cite{Kibble_1945, mehler1866ueber}, we get
        \begin{align}
            p(a, b) = p(a) p(b) \sum_{l=0}^{\infty}\frac{(\rho/c)^l}{l!} H_l(a) H_l(b).
        \end{align}
        Then, it follows
        \begin{align}
            \mathbb{E}_{z_1, z_2}&[H_i(\rho z_1 +  \sqrt{c^2- \rho^2} z_2) H_j(z_1)] \\
            &= \mathbb{E}_{a,b} [H_i(c a) H_j(b)],\\
            &= \int_{-\infty}^{\infty} \int_{-\infty}^{\infty} H_i(c a) H_j(b) p(a,b) d a d b,\\
            &=\int_{-\infty}^{\infty} \int_{-\infty}^{\infty} H_i(c a) H_j(b) p(a) p(b) \sum_{l=0}^{\infty}\frac{(\rho/c)^l}{l!} H_l(a) H_l(b) d a d b,\\
            &= \sum_{l=0}^{\infty}\frac{(\rho/c)^l}{l!}\int_{-\infty}^{\infty} H_i(c a) H_l(a) p(a) d a\int_{-\infty}^{\infty}  H_j(b) H_l(b) p(b)  d b,\\
            &= \sum_{l=0}^{\infty}\frac{(\rho/c)^l}{l!} \E_a[ H_i(c a) H_l(a)] \E_b [H_j(b) H_l(b)],\\
            &= \sum_{l=0}^{\infty}\frac{(\rho/c)^l}{l!} (i!) (j!) (c^i) \delta_{i,l} \delta_{j,l},\\
            &= (i!) \rho^i \delta_{i j},
        \end{align}
        where we use Lemma \ref{supplement_lemma:hermite_orthogonality} to reach the line before the last one.
    \end{proof}
    \label{supplement_lemma:hermite_correlated}
\end{lemma}

\begin{lemma}
    Let $H_i(x)$ be the probabilist's $i$-th Hermite polynomial. Also, let $z \sim \Normal(0,1)$. Then, for any $c \in \R$,
    \begin{equation}
            \E [H_i(c z) H_j(z)] = (i!) (c^i) \delta_{i j}. 
            \label{supplement_eq:hermite_orthogonality}
    \end{equation}
    \begin{proof}
        It is known that $\E[H_i(z) H_j(z)] = (i!) \delta_{i j}$ \cite[Chapter 11.2]{O’Donnell_2014}. Here, we extend it to our case. For $i \neq j$, $\E [H_i(c z) H_j(z)] = 0$ by orthogonality. Let $a_l$ be the coefficient of $x^l$ in $H_i(x)$. Then, we have
        \begin{align}
            \E [H_i(c z) H_i(z)] &= \sum_{l=0}^{i} c^l  \E[a_l z^l H_i(z)],\\
            &\stackrel{(a)}{=} c^i \E[a_l z^i H_i(z)],\\
            &\stackrel{(b)}{=} c^i \E[H_i(z) H_i(z)],\\
            &\stackrel{(c)}{=} (i!) (c^i),
        \end{align}
        where we use $\E[z^l H_i(z)] = \delta_{l j}$ due to orthogonality to reach (a) and (b) while (c) is because of $\E[H_i(z) H_i(z)] = (i!)$. Combining $i = j$ and $i \neq j$ cases, we get \eqref{supplement_eq:hermite_orthogonality}.
    \end{proof}
    \label{supplement_lemma:hermite_orthogonality}
\end{lemma}

\section{Plots for Training Errors}
For the sake of completeness, here, we provide the supplementary training error plots corresponding to the setting of Figure \ref{fig:linear_equivalence} and Figure \ref{fig:polynomial_equivalence} in Section \ref{sec:main_results}.
\begin{figure}
    \centering
     \includegraphics[width=0.35\linewidth]{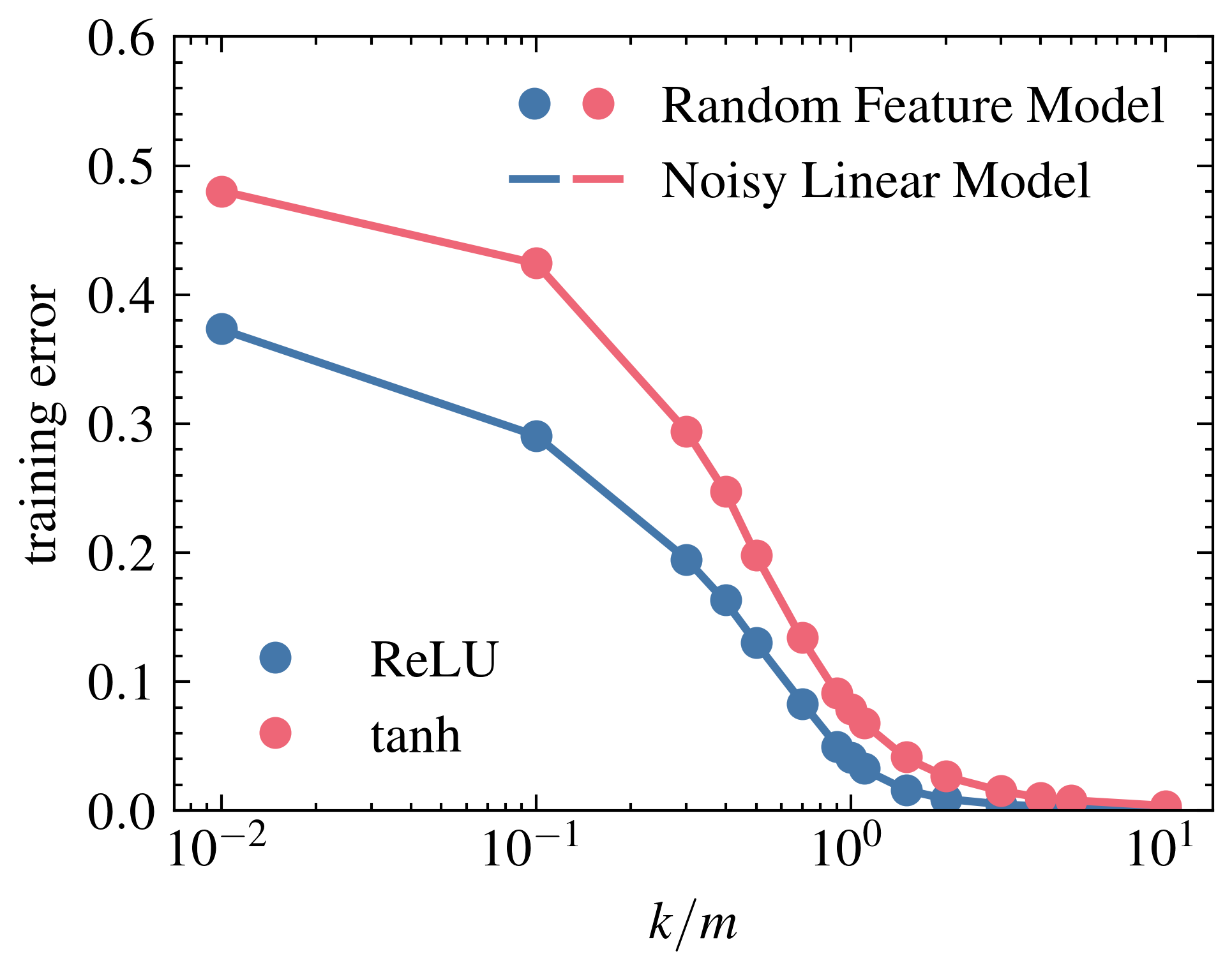}
     \captionsetup{justification=centering}
     \caption{Training errors for the misaligned case (the setting in Figure \ref{fig:linear_equivalence})}
\end{figure}

\begin{figure}[H]
    \centering
    \begin{subfigure}[b]{0.35\textwidth}
         \centering
         \includegraphics[width=0.9\linewidth]{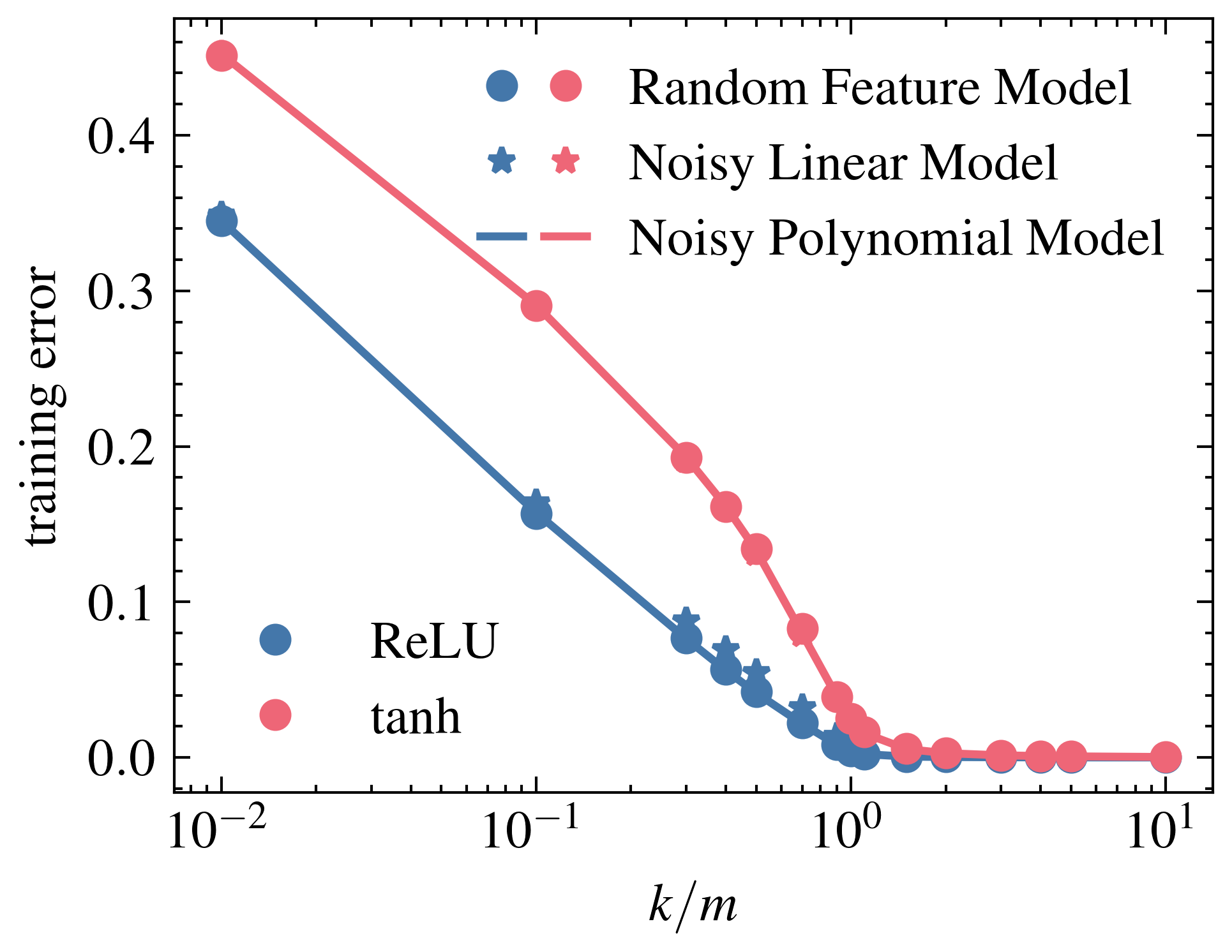}
         \captionsetup{justification=centering}
         \caption{$\sigma_* = \sigma_{ReLU}$}
     \end{subfigure}
     \begin{subfigure}[b]{0.35\textwidth}
         \centering
         \includegraphics[width=0.9\linewidth]{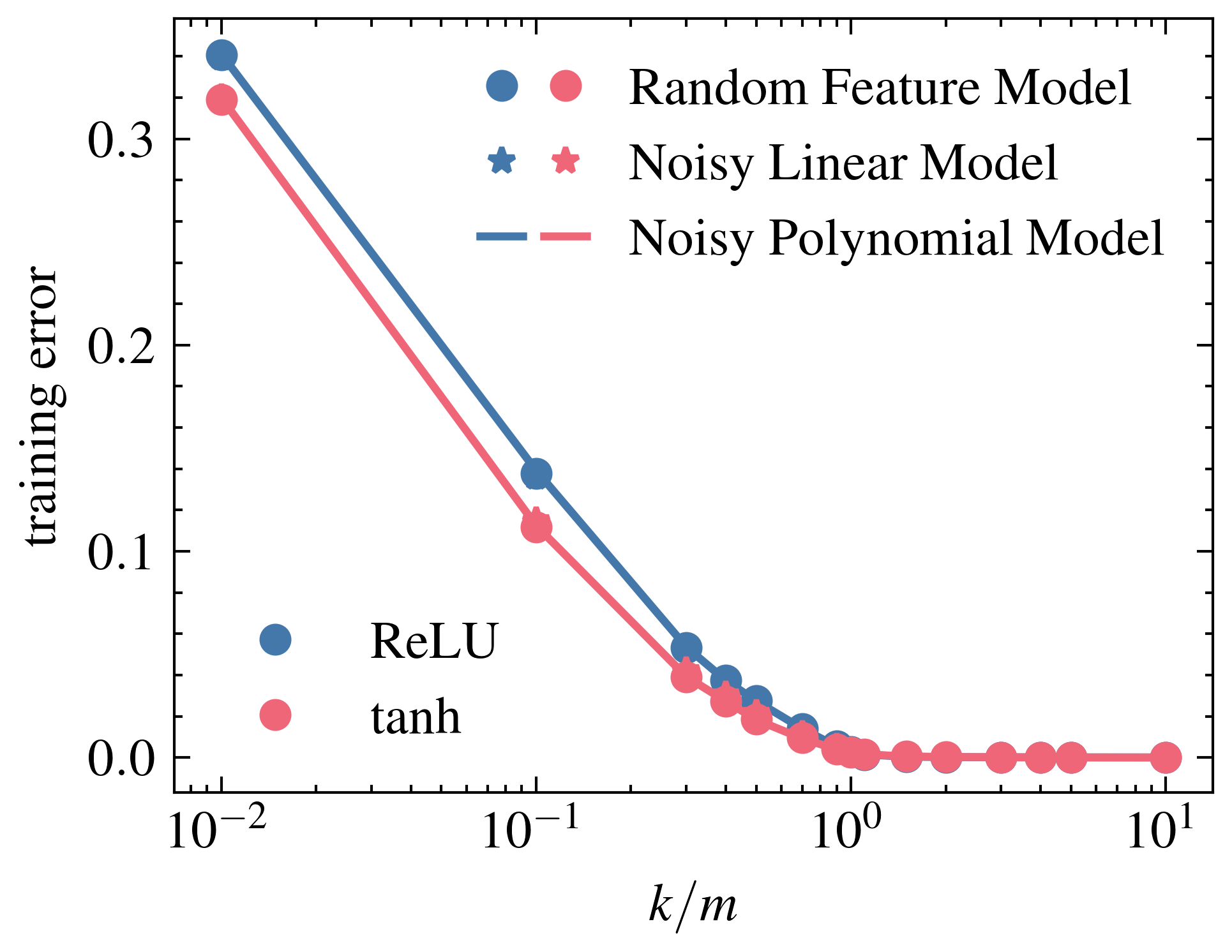}
         \captionsetup{justification=centering}
         \caption{$\sigma_* = \sigma_{tanh}$}
     \end{subfigure}
     \caption{Training errors for the aligned case (the setting in Figure \ref{fig:polynomial_equivalence}).}
     \label{supplement_fig:training_plots}
\end{figure}

\begin{figure}
    \centering
    \begin{subfigure}[b]{0.35\textwidth}
         \centering
         \includegraphics[width=0.9\linewidth]{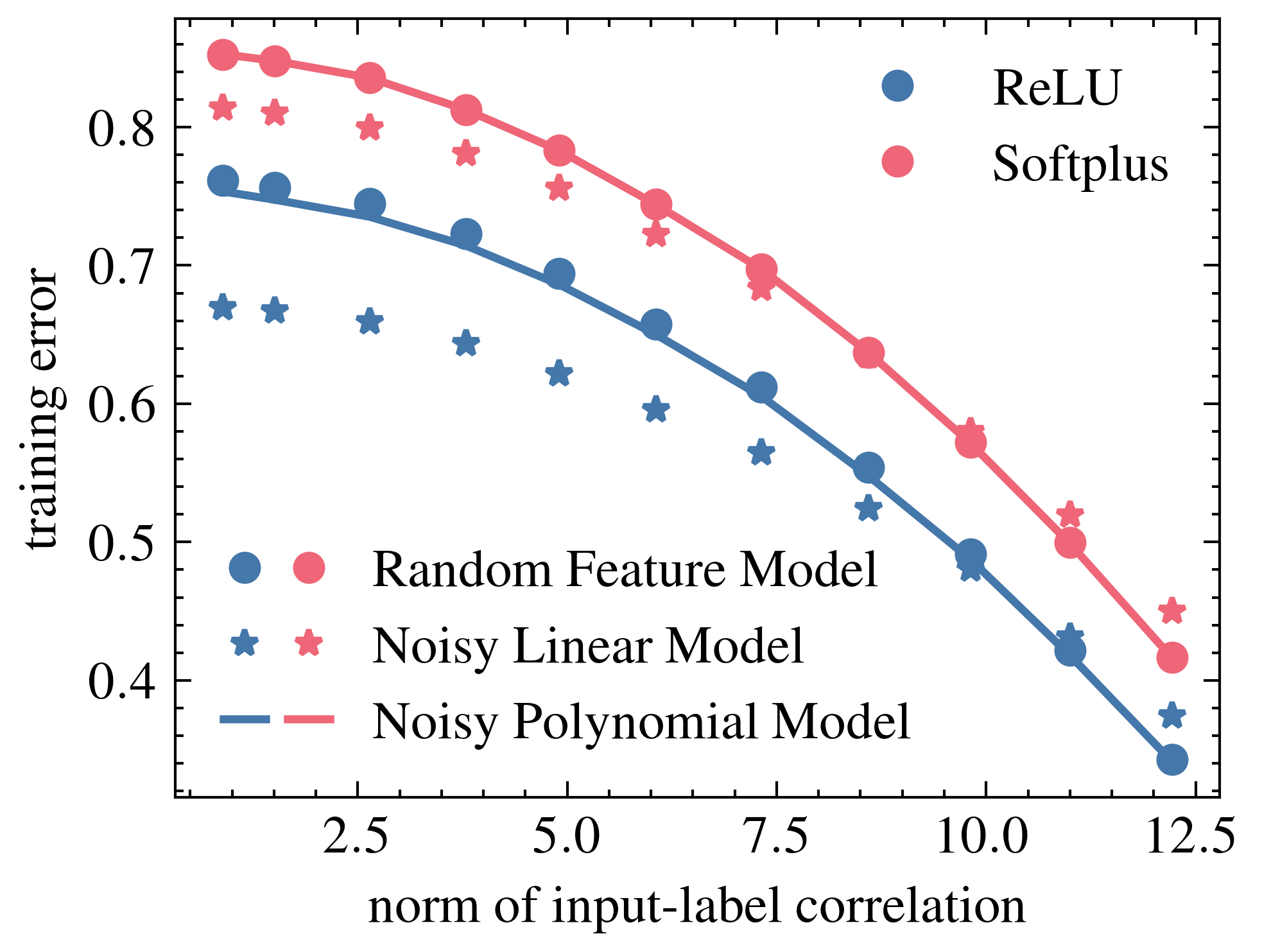}
         \captionsetup{justification=centering}
         \caption{Original Inputs}
     \end{subfigure}
     \begin{subfigure}[b]{0.35\textwidth}
         \centering
         \includegraphics[width=0.9\linewidth]{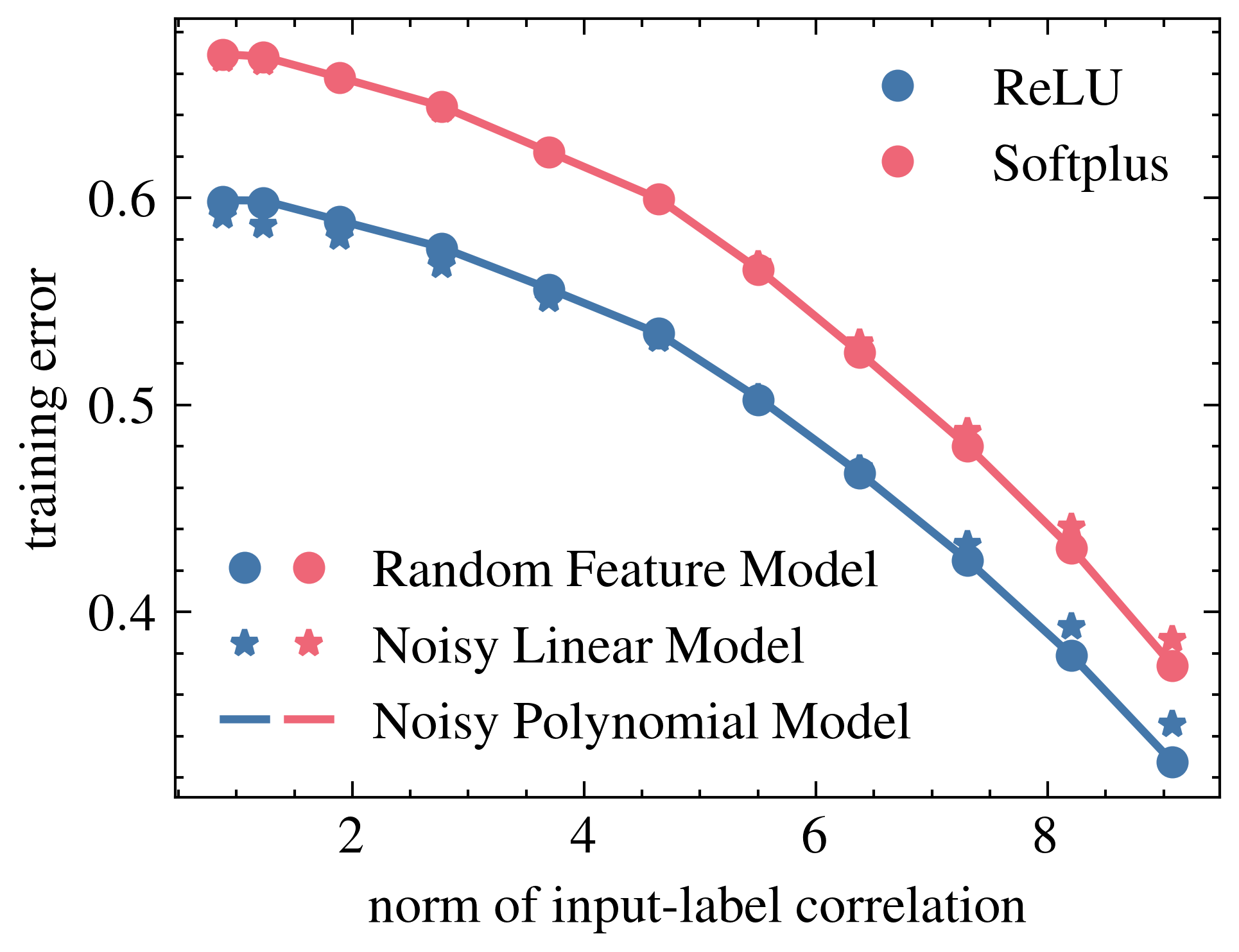}
         \captionsetup{justification=centering}
         \caption{Inputs with Gaussian Noise}
     \end{subfigure}
     \caption{Training errors for CIFAR-10 experiments (the setting in Figure \ref{fig:CIFAR-10}).}
     \label{supplement_fig:training_plots_CIFAR-10}
\end{figure}

\section{Details for CIFAR-10 Experiments}
\label{supplement:CIFAR-10}
In our experiments, we focus on binary classification between airplanes and automobiles using the CIFAR-10 dataset \cite{krizhevsky2009learning} to demonstrate how our findings translate to real-world applications. We randomly select 2,000 samples from each class for training, while a separate set of 2,500 samples (distinct from the training set) is used to calculate the test (generalization) error. To prepare the input samples, we normalize the pixel values to achieve zero mean and unit variance for each color channel (R, G, B). Specifically, the pixel values are first scaled to the range [0, 1] by dividing by 255, followed by subtracting channel-wise means and dividing by their respective standard deviations. The images are then flattened into vectors for input into the model.

For the feature matrix $\mathbf{F}$, we sample entries independently from a normal distribution $\mathcal{N}(0, 1/\text{Tr}(\mathbb{E}[\mathbf{x} \mathbf{x}^T]))$, ensuring that $\mathbb{E}[(\mathbf{f}_i^T \mathbf{x})^2] = 1$ for all $i$, where $\mathbf{f}_i$ denotes the $i$-th row of $\mathbf{F}$. 

Our results are presented in two scenarios: first with the original normalized inputs (Figure \ref{fig:CIFAR-10}a), and second after adding standard Gaussian noise (variance of one for each feature) to create a more isotropic covariance structure (Figure \ref{fig:CIFAR-10}b). This addition of noise allows us to assess the performance equivalence between the linear model and RFM under conditions of weak input-label correlation.

For labels, we use $\{-1,1\}$ for the first class and the second class, respectively. To control input-label correlation, we introduce a label flipping mechanism with probability $p$, where $p=0$ corresponds to true labels (maximum correlation) and $p=0.5$ represents random labels (minimum correlation). By varying $p$ within the range $[0, 0.5]$, we interpolate between these two extremes and analyze how this affects model performance. The training errors for these experiments are illustrated in Figure \ref{supplement_fig:training_plots_CIFAR-10}.

\end{document}